\documentclass[10pt,twocolumn,letterpaper]{article}

\usepackage[pagenumbers]{cvpr} %

\usepackage{duckuments}

\usepackage{multirow}
\usepackage{pifont}
\newcommand{\cmark}{\ding{51}\xspace}

\newcommand{\xmark}{\ding{55}\xspace}
\usepackage{makecell}
\usepackage{etoc}
\usepackage{titletoc}
\usepackage{color, colortbl}
\usepackage[dvipsnames]{xcolor}
\usepackage{subcaption}
\usepackage{graphbox}
\usepackage{tcolorbox}
\tcbuselibrary{skins,breakable}
\usepackage{longtable}

\usepackage{epigraph}
\newcommand{\red}[1]{{\color{red}#1}}

\newcommand{\nusdata}{nuScenes\xspace}

\newcommand{\datasuitename}{OpenDV-2K\xspace}
\newcommand{\youtubesplitename}{OpenDV-YouTube\xspace}
\newcommand{\splitname}[1]{OpenDV-#1\xspace}

\newcommand{\modelname}{GenAD\xspace}

\definecolor{Gray}{gray}{0.9}
\definecolor{LightCobaltBlue}{RGB}{143,170,220}
\definecolor{Amber}{RGB}{255,102,0}
\definecolor{BlackOlive}{RGB}{59,56,56}
\definecolor{AlloyOrange}{RGB}{197,90,17}
\definecolor{B'dazzledBlue}{RGB}{47,85,151}
\definecolor{DarkBlue}{RGB}{72, 116, 203}
\definecolor{DarkGreen}{RGB}{106, 169, 63}
\newcommand{\cmtt}[1]{{\fontfamily{cmtt}\selectfont #1}}

\newcommand{\sdxl}{SDXL\xspace}
\newcommand{\unet}{UNet\xspace}
\newcommand{\gauss}{Gaussian\xspace}

\usepackage{amsmath,amsfonts,bm,mathtools}

\def\eqref#1{equation~\ref{#1}}

\def\1{\bm{1}}

\def\rvepsilon{{\boldsymbol{\epsilon}}}

\def\rvc{{\mathbf{c}}}

\def\rvf{{\mathbf{f}}}

\def\rvv{{\mathbf{v}}}

\def\rvx{{\mathbf{x}}}

\DeclareMathAlphabet{\mathsfit}{\encodingdefault}{\sfdefault}{m}{sl}
\SetMathAlphabet{\mathsfit}{bold}{\encodingdefault}{\sfdefault}{bx}{n}

\definecolor{cvprblue}{rgb}{0.21,0.49,0.74}
\usepackage[pagebackref,breaklinks,colorlinks,citecolor=cvprblue]{hyperref}

\usepackage[capitalize]{cleveref}
\crefname{section}{Sec.}{Secs.}
\Crefname{section}{Section}{Sections}
\Crefname{table}{Table}{Tables}
\crefname{table}{Tab.}{Tabs.}

\title{Generalized Predictive Model for Autonomous Driving
}

\author{
{Jiazhi Yang}$^{1\ast}$ \quad
{Shenyuan Gao}$^{2,1\ast}$ \quad
{Yihang Qiu}$^{1\ast}$ \quad
{Li Chen}$^{3,1\dagger}$ \quad
{Tianyu Li}$^{1}$ \quad
{Bo Dai}$^{1}$ \\
{Kashyap Chitta}$^{4,5}$ \quad
{Penghao Wu}$^{1}$ \quad
{Jia Zeng}$^{1}$ \quad
{Ping Luo}$^{3}$ \quad
{Jun Zhang}$^{2\natural}$ \\
{Andreas Geiger}$^{4,5\natural}$ \quad
{Yu Qiao}$^{1\natural}$ \quad
{Hongyang Li}$^{1\dagger}$ \\
[2mm]
$^1$~OpenDriveLab and Shanghai AI Lab  
\quad
$^2$~Hong Kong University of Science and Technology \\
$^3$~University of Hong Kong \quad
$^4$~University of Tübingen \quad
$^5$~Tübingen AI Center
}

\begin{document}

\twocolumn[{%
\renewcommand\twocolumn[1][]{#1}%
\maketitle
\begin{center}
    \centering
    \captionsetup{type=figure}
    \includegraphics[width=\linewidth]{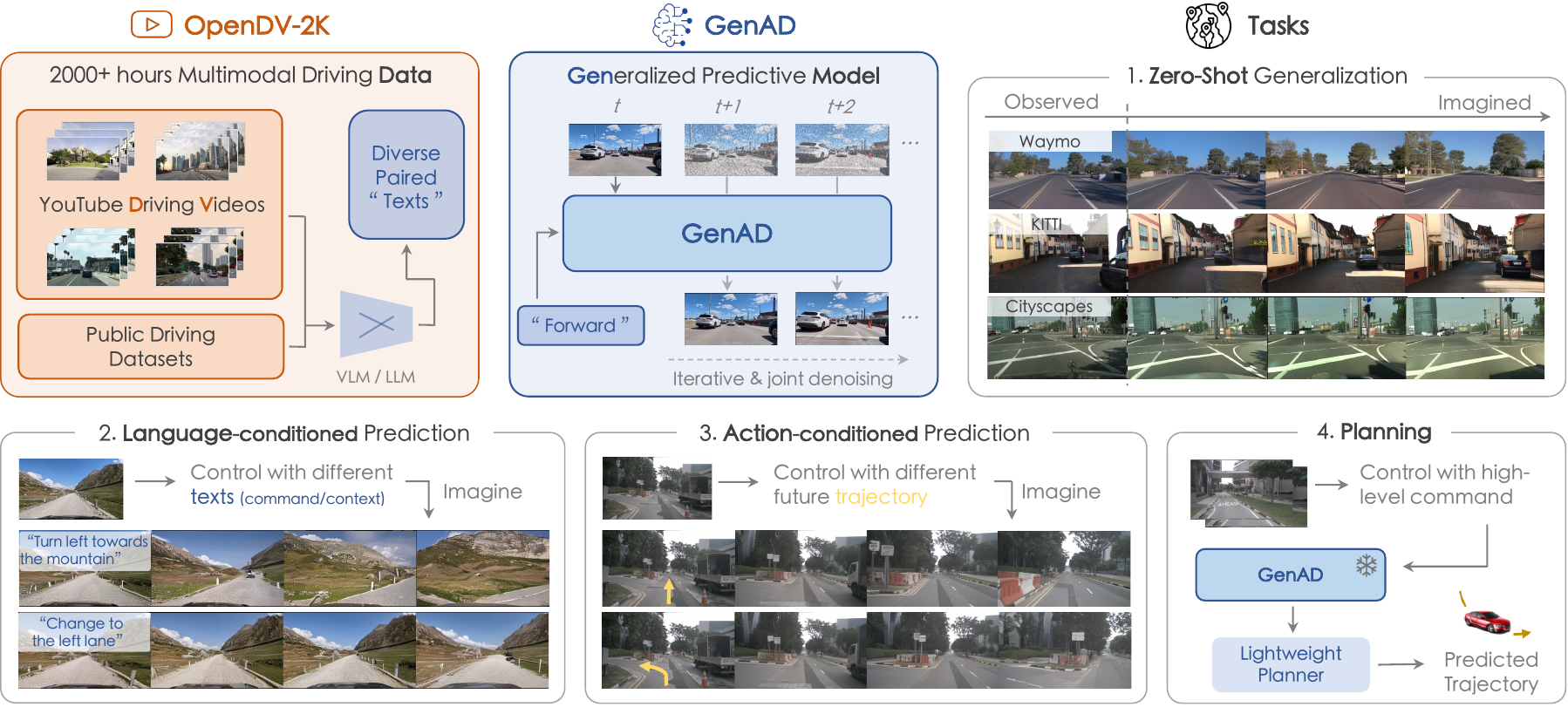} 
    \vspace{-20pt}
     \captionof{figure}{\label{fig:teaser}
    \textbf{Overview of the \modelname paradigm}. 
    We aim to establish a generalized video prediction 
    paradigm
    for autonomous driving by presenting the largest  
    multimodal 
    driving
    video dataset to date, \textcolor{AlloyOrange}{\textbf{\datasuitename}}, and a generative 
    model that predicts the future 
    given past visual 
    and textual input,
    \textcolor{B'dazzledBlue}{\textbf{\modelname}}. The strong generalization 
    and controllability of \modelname is validated spanning a diverse spectrum of tasks,
    including
    zero-shot domain transfer,
    language-conditioned prediction, action-conditioned prediction, and motion planning. 
    }
\end{center}}]

{\let\thefootnote \relax \footnote{$^*$Equal contribution, ordered by coin toss. 
$^\natural$Equal co-advising.}
\let\thefootnote \relax \footnote{
$^\dagger$Project lead. Primary contact:
\cmtt{yangjiazhi@opendrivelab.com}
}}

\setcounter{footnote}{0} 

\begin{abstract}
In this paper, we introduce the first large-scale video prediction model in the autonomous driving discipline. To eliminate the restriction of high-cost
data collection and empower the generalization ability of our model, we acquire
massive data from the web and pair it with diverse and high-quality text descriptions. 
The resultant dataset accumulates over 2000 hours of driving videos, spanning areas all over the world with diverse weather conditions and traffic scenarios.
Inheriting the merits from recent latent diffusion models, our model, dubbed \modelname, handles the challenging dynamics in driving scenes with novel temporal reasoning blocks.
We showcase that it can generalize to various unseen driving datasets in a zero-shot manner, surpassing 
general or 
driving-specific
video prediction counterparts. 
Furthermore, 
\modelname
can be adapted into an action-conditioned prediction model or a motion planner, 
holding great potential for real-world driving applications.
\end{abstract}

\section{Introduction}
\label{sec:intro}

Autonomous driving agents, as a promising application of high-level artificial intelligence, perceive the surrounding environment, build internal world model representations, make decisions, and take actions in response~\cite{Chen2023E2ESurvey,Xi2023LLMAgentSurvey}. However, despite dedicated efforts in academia and industry for decades, their deployment is still restricted to certain areas or scenarios, and they cannot be applied over the world seamlessly. 
One critical reason is the limited generalization ability of learned models in structured autonomous driving systems. Typically, perception models face challenges of generalizing to diverse environments with changes in geographical locations, sensor configurations, weather conditions, open-set objects, \etc; prediction and planning models fail to generalize to nondeterministic futures with rare scenarios and different driving intentions~\cite{Zhu2023AnyD, Bahari2022CVPR, Hanselmann2022KING}.

Motivated by how humans learn to perceive and cognize the world~\cite{Wolpert1995InternalModel, Lecun2015DL,lecun2022PATH},
we advocate employing driving videos as the universal interface that generalizes to diverse environments with dynamic futures. Based on this, a driving video predictive model is preferred to fully capture the world knowledge about driving scenarios (\cref{fig:teaser}). 
By predicting the future, 
the video predictor 
essentially
learns two vital aspects of autonomous driving: how the world operates, and how to maneuver safely in the wild.

Recently, the community has begun to adopt video as the interface to represent observation behavior and action 
for various robot tasks~\cite{Dai2023UniPi}. 
For domains such as classical video prediction and robotics,
the video backgrounds are mostly static, the movement of robots is slow, and the resolution of videos is low. 
In contrast, for the driving scenarios,
it struggles with outdoor environments being highly dynamic, agents encompassing much larger motions, and the sensory resolution covering a large range of view. 
These distinctions lead to substantial challenges for 
autonomous driving applications. 
Fortunately, there are some preliminary attempts on
developing a video predictive model in the driving domain~\cite{Qi20193DMotion, Kim2021DriveGAN, Blattmann2023VideoLDM, Gupta2023MaskViT, Hu2023GAIA, Wang2023DriveDreamer, Wang2023DriveWM, Jia2023Adriver-1, Lu2023Wovogen}.
Despite promising progress in terms of prediction quality, 
these 
attempts
have not achieved 
desirable capability of
generalization 
as in classical robot tasks (\eg, manipulation), being 
confined
to either limited scenarios such as highways with low traffic density~\cite{Blattmann2023VideoLDM} and small-scale datasets~\cite{Gupta2023MaskViT, Wang2023DriveWM, Lu2023Wovogen, Jia2023Adriver-1, Wang2023DriveDreamer}, or restricted conditions that raises difficulties to generate diverse environments~\cite{Qi20193DMotion}.
How to unveil the potential of video prediction models for driving remains seldom explored.

Motivated by the discussions above,
we target at building a video predictive model for autonomous driving, 
capable of generalizing to new conditions and environments.
To this end, we have to answer the following
questions: \textit{(1) What 
data can be obtained in a feasible and scalable manner? (2) How can we formulate a predictive model to capture the complex evolution of dynamic scenarios? (3) How can we apply the (foundation) model for downstream tasks?}

\smallskip
\noindent\textbf{Scaled Data.}
To achieve powerful generalization ability, a substantial and diverse corpus of data is 
necessary.
Inspired by the success of learning from Internet-scale data in foundation models~\cite{Radford2021CLIP, Alayrac2022Flamingo, Kirillov2023SAM}, we construct our driving dataset from both the web and publicly licensed datasets.
Compared to existing options,
which are limited in scale and diversity due to their regulated collection processes, online data owns great diversity in several aspects: geographic locations, terrains, weather conditions, safety-critical scenarios, sensor settings, traffic elements, \etc.
To 
guarantee
the data is 
of
high-quality and desirable 
for large-scale training, we exhaustively collect driving recordings on YouTube and remove unintended corruption frames via 
rigorous human verification.
Furthermore, 
videos are paired with diverse text-level conditions, including descriptions generated and refined with the aid of existing foundation models~\cite{Li2023BLIP2, OpenAI2022GPT4},
and high-level instructions inferred by a video classifier.
Through these steps, we construct \textbf{\datasuitename}, the \textit{largest} public driving dataset to date, containing more than 2000 hours of driving videos and being 374 times larger than the widely used \nusdata counterpart.
Our dataset is publicly available at \url{https://github.com/OpenDriveLab/DriveAGI}.

\begin{figure*}[h!]
\begin{minipage}[b]{0.67\textwidth}
    \centering
    \footnotesize
    \begin{tabular}{c|c|cc|cc|c}
        \toprule
        \multirow{2}{*}{} 
        & \multirow{2}{*}{Dataset} 
        & \multirow{2}{*}{\makecell{Duration\\(hours)}} 
        & \multirow{2}{*}{\makecell{Front-view\\Frames}} 
        & \multicolumn{2}{c|}{Geographic Diversity} & \multirow{2}{*}{\makecell{Sensor\\Setup}} \\
        & & & & Countries & Cities &\\
        \midrule
        \xmark & KITTI~\cite{Geiger2013KITTI} & 1.4 & 15k & 1 & 1 & fixed\\
        \xmark & Cityscapes~\cite{Cordts2016Cityscapes} & 0.5 & 25k & 3 & 50 & fixed \\
        \xmark & Waymo Open$^\star$~\cite{Sun2019WOD} & 11 & 390k & 1 & 3 & fixed\\
        \xmark & Argoverse 2$^\star$~\cite{Wilson2023Argoverse2} & 4.2 & 300k & 1 & 6 & fixed\\
        \midrule
        \cmark & \nusdata~\cite{Caesar2019nuScenes} & 5.5 & 241k & 2 & 2 & fixed\\
        \cmark & nuPlan$^\star$~\cite{Caesar2021nuPlan} & 120 & 4.0M & 2 & 4 & fixed\\
        \cmark & Talk2Car~\cite{Deruyttere2019Talk2Car} & 4.7 & - & 2 & 2 & fixed\\
        \cmark & ONCE~\cite{Mao2021ONCE} & 144 & 7M & 1 & - & fixed\\
        \cmark & Honda-HAD~\cite{Kim2019HondaHAD} & 32 & 1.2M & 1 & - & fixed \\
        \cmark & Honda-HDD-Action~\cite{Ramanishka2018HondaHDD} & 104 & 1.1M & 1 & - & fixed \\
        \cmark & Honda-HDD-Cause~\cite{Ramanishka2018HondaHDD} & 32 & - & 1 & - & fixed\\
        \midrule
        \cmark & \youtubesplitename (Ours) & 1747 & 60.2M & {$\geq$40}$^\dagger$ & {$\geq$244}$^\dagger$ & {uncalibrated}\\
        - & \textbf{\datasuitename} (Ours) & \textbf{2059} & \textbf{65.1M} & \textbf{$\geq$40}$^\dagger$ & \textbf{$\geq$244}$^\dagger$ & \textbf{uncalibrated}\\
        \bottomrule
    \end{tabular}
    \vspace{-6pt}
    \captionof{table}{\textbf{
    \datasuitename comparison at a glance to existing counterparts 
    in terms of scale and diversity}.
    Note that datasets with \cmark are included in \datasuitename (last row). 
    $^\star$Perception 
    subset
    in Waymo Open, Argoverse 2, and nuPlan. 
    $^\dagger$Estimated by GPT~\cite{Ouyang2022InstructGPT} from video titles.
    }
    \label{tab:dataset_compare}
\end{minipage}
\hfill
\begin{minipage}[b]{0.31\textwidth}
    \centering
    \includegraphics[width=\textwidth,align=b]{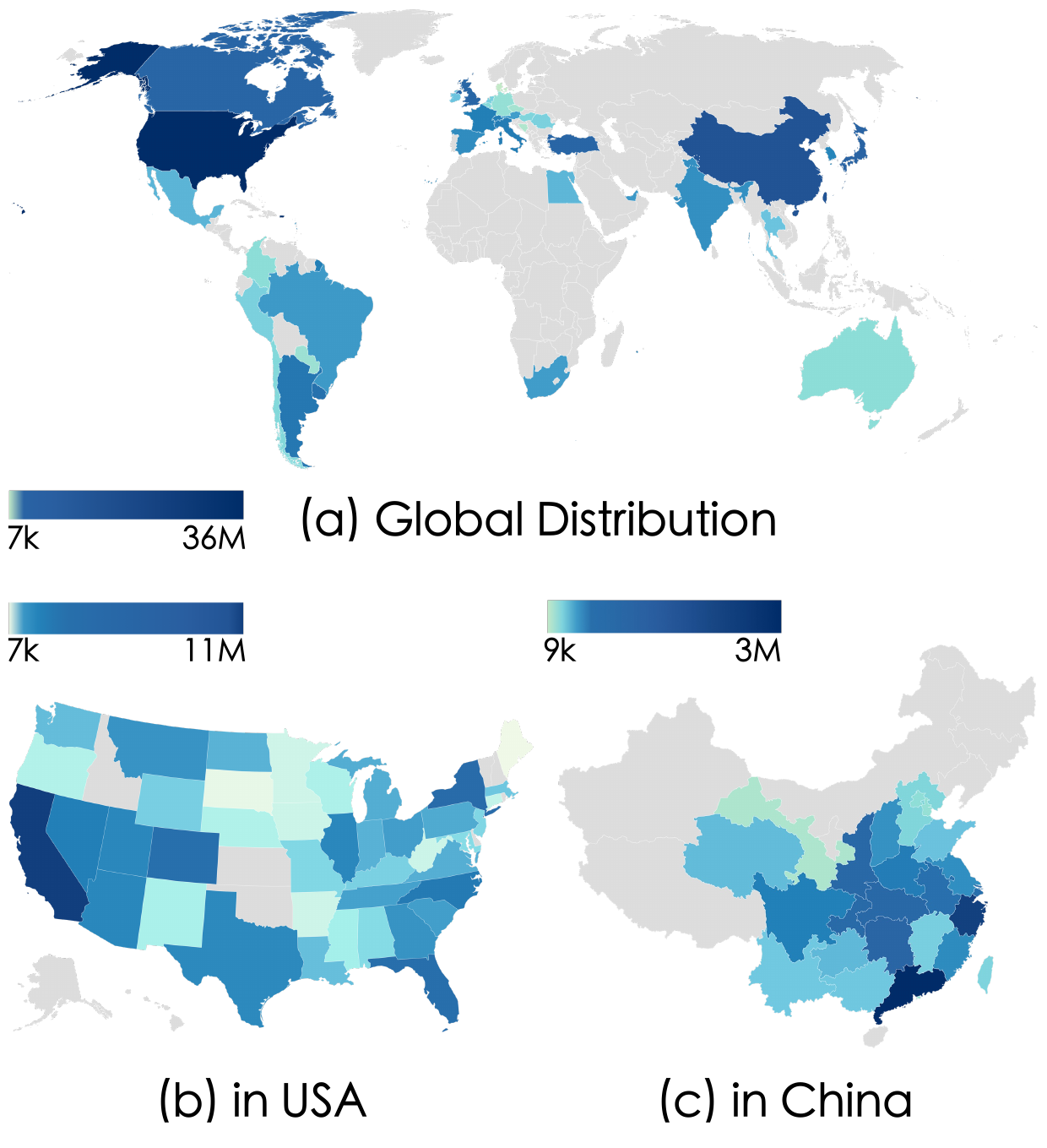}
    \vspace{-20pt}
    \captionof{figure}{\textbf{Geographic distribution of \datasuitename}.
    Our dataset covers ample driving scenarios
    around  
    the world.
    }
    \label{fig:dataset-diversity}
\end{minipage}
\vspace{-.1cm}
\end{figure*}

\smallskip
\noindent\textbf{Generalized Predictive Model.}
Learning a generalized driving video predictor 
bears
several key challenges: generation quality, training efficiency, causal reasoning, and drastic view shift. We address these aspects 
by 
presenting
a novel temporal generative model with two-stage learning.
To capture the environment details, enhance generation quality, and maintain training efficiency simultaneously, we build upon the recent success of \textit{latent diffusion models} (LDMs)~\cite{Rombach2022LDM, Podell2023SDXL}. 
In the first stage,
we transfer the generation distribution of 
LDM from its pre-trained general vision domain to the driving domain by fine-tuning it on \datasuitename images. 
In the second stage, 
we interleave the proposed temporal reasoning blocks into the original model and learn to predict the future given past frames and conditions.
Contrary to conventional temporal modules~\cite{Blattmann2023VideoLDM, Ho2022VideoDiffusionModels} that suffer from causal confusion and large motion, our 
solution
consists
of causal temporal attention and decoupled spatial attention to efficiently model the drastic spatiotemporal 
shift
in highly dynamic driving scenes.
After sufficient training, our \textbf{Gen}erative model for \textbf{A}utonomous \textbf{D}riving (\textbf{\modelname})\footnote{
Note that \modelname is abbreviated from both \textbf{Gen}erative models and \textbf{Gen}eralized capabilities.
} can generalize to various 
scenarios in a zero-shot fashion.

\smallskip
\noindent\textbf{Extensions for Simulation and Planning.} 
After large-scale pre-training of video prediction, \modelname essentially understands how the world evolves and how to drive. We show how to adapt its learned knowledge for real-world driving problems, \ie, simulation and planning. For simulation, we fine-tune the pre-trained model with future ego trajectories as additional conditions, to associate future imaginations with different ego actions. 
We also empower \modelname to perform planning on challenging benchmarks by using a lightweight planner to translate latent features into the future trajectory of the ego vehicle.
On account of its pre-trained ability to predict accurate future frames, our algorithm exhibits promising results in both simulation consistency and planning reliability.

\section{\datasuitename Dataset}
\label{sec:dataset}

We introduce \datasuitename, a large-scale multimodal dataset for autonomous driving, to support the training of a generalized video prediction model. The main component is a vast corpus of high-quality YouTube driving videos, which are collected from all over the world, and are gathered into our dataset after a careful curation process.
We automatically create language annotations for these videos using vision-language models.
To further improve its diversity in sensor configurations and language expressions, we merge 7 publicly licensed datasets into our \datasuitename, as shown in \cref{tab:dataset_compare}.
As a result, \datasuitename 
occupies
a total of 2059 hours of videos paired with texts, including 1747 hours from YouTube and 312 hours from public datasets. 
We use \youtubesplitename and \datasuitename to specify the YouTube split and the overall dataset, respectively.

\subsection{Diversity over Prior Datasets}
A brief comparison with other public datasets is provided in \cref{tab:dataset_compare}. Beyond its significant scale, the proposed \datasuitename 
represents
\textit{diversity} 
across various 
aspects as follows.

\smallskip
\noindent\textbf{Globe-wise Geographic Distribution.} 
Due to the global nature of online videos, \datasuitename covers 
more than 40 countries and 244 cities worldwide. This is a tremendous improvement over previous public datasets, which are typically gathered in a small number of restricted areas. We plot the specific distribution of \youtubesplitename in \cref{fig:dataset-diversity}. 

\smallskip
\noindent\textbf{Open-world Driving Scenarios.} 
Our dataset provides a huge amount of realistic driving experience in the open world, 
covering rare environments like forests, extreme weather conditions like heavy snow, and appropriate driving behaviors in response to interactive traffic situations.
These data are crucial for diversity and generalization yet are seldom collected in existing public datasets.

\smallskip
\noindent\textbf{Unrestricted Sensor Configurations.}
Current driving datasets are confined to specific sensor configurations, including intrinsic and extrinsic camera parameters, image, sensor type, optics, \etc, which poses great challenges for deploying the learned models with different sensors~\cite{Li2022STMono3D}. 
In contrast, YouTube driving videos are recorded in various types of vehicles with flexible camera setups, which aids in the robustness of the trained model when deployed using a novel camera setting.

\begin{figure}[t]
    \centering
    \includegraphics[width=0.9\linewidth]{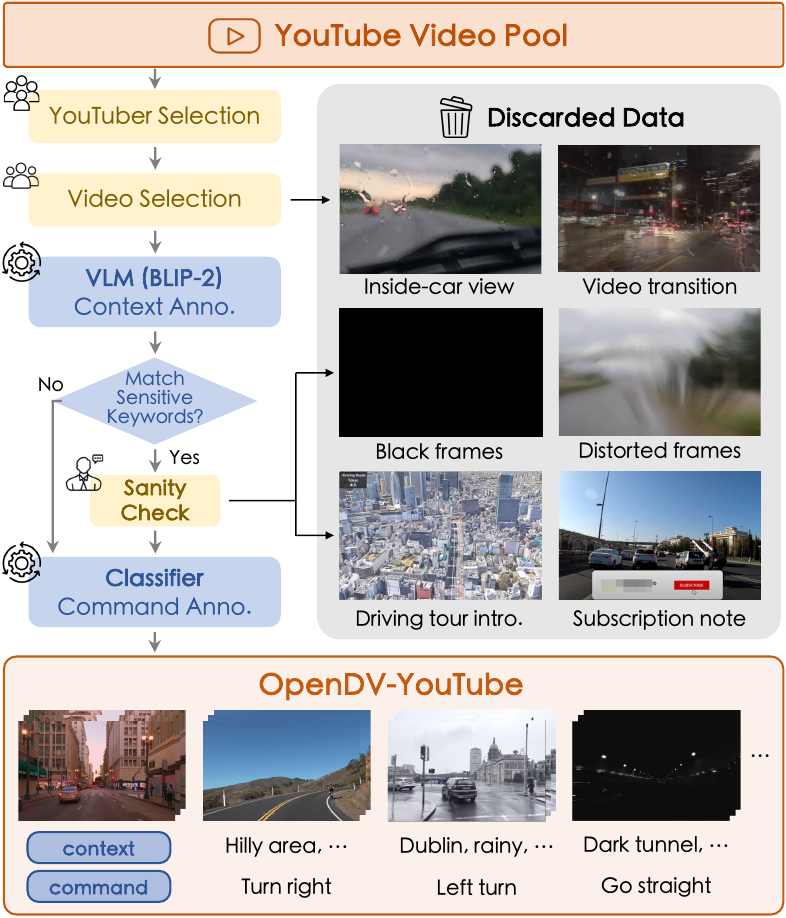}
    \vspace{-6pt}
    \caption{\textbf{Dataset construction of \youtubesplitename with quality check in the loop.}
    We collect videos from YouTubers with 
    qualified 
    driving videos, and dispose of those 
    with inappropriate viewpoints or involving scene transitions. 
    Then each frame is described with language contexts using VLM 
    followed by keyword checks on texts,
    such as ``words", ``watermark", ``dark", ``blurry", \etc. Through this process, distorted or 
    entirely
    black images are wiped out. A classifier tags videos with high-level intentions as commands, 
    incubating
    the final 
    data
    corpus
    of high-quality video-text pairs being 1747 hours long.}
    \label{fig:dataset-construction}
    \vspace{-.3cm}
\end{figure}

\subsection{Towards High-quality Multimodal Dataset}
\noindent\textbf{Driving Video Collection and Curation.}
Finding clean driving videos from the vast pool of the web is a tedious and costly task. To simplify the process, we start by selecting certain video uploaders, \ie, YouTubers.
Judging from the average length and overall quality, we collect 43 YouTubers with 2139 high-quality front-view driving videos. 
To make sure there is no overlap between training and validation sets, we take all videos from 3 YouTubers for validation, with the remaining videos as the training set.
To rule out non-driving frames like video introductions and subscription reminders, we discard a certain length of segments at the beginning and end of each video. 
Each frame is then described with language contexts using a VLM model, BLIP-2~\cite{Li2023BLIP2}.
We further remove the black frames and transition frames, which are not ideal for training, by manually checking if there are certain keywords in these contexts. An illustration of the dataset construction pipeline is in \cref{fig:dataset-construction}, and we introduce how to generate the contexts below.

\smallskip
\noindent\textbf{Language Annotation for YouTube Videos.}
To create a predictive model that can be controlled by natural language to simulate different futures accordingly,
To make the predictive model controllable and improve the sample quality~\cite{bao2022SCDM}, it is crucial to pair the driving videos with meaningful and varied language annotations. 
We construct two types of texts for \youtubesplitename, \ie, driving commands for ego-vehicle and frame descriptions, namely ``command" and ``context", to help the model comprehend ego actions and open-world concepts, respectively. 
For commands, we train a video classifier on Honda-HDD-Action~\cite{Ramanishka2018HondaHDD} for 14 types of actions to label ego behaviors in a 4s sequence. These categorical commands will be further mapped to multiple free-form expressions from a predefined dictionary. 
For contexts, we leverage an established vision-language model, BLIP-2~\cite{Li2023BLIP2}, to describe the main objects and scenarios for each frame.
For more details on annotations, please refer to \cref{supp:anno}.

\smallskip
\noindent\textbf{Enlarging Language Spectrum with Public Datasets.}
Considering that BLIP-2 annotations are generated for static frames without comprehension of dynamic driving scenarios such as the traffic light transitions, we exploit several public datasets that provide linguistic descriptions for driving scenarios~\cite{Caesar2019nuScenes, Caesar2021nuPlan, Deruyttere2019Talk2Car, Mao2021ONCE, Kim2019HondaHAD, Ramanishka2018HondaHDD}. 
However, their metadata is relatively sparse with only a few words such as ``sunny road''. We further enhance their text quality using GPT~\cite{Ouyang2022InstructGPT} to form a descriptive ``context" and generate a ``command" by categorizing the logged trajectory for each video clip.
Ultimately, we integrate these datasets with \youtubesplitename to establish \datasuitename dataset, as shown in the last row of \cref{tab:dataset_compare}.

\begin{figure*}[t!]
    \centering
    \includegraphics[width=1.\textwidth]{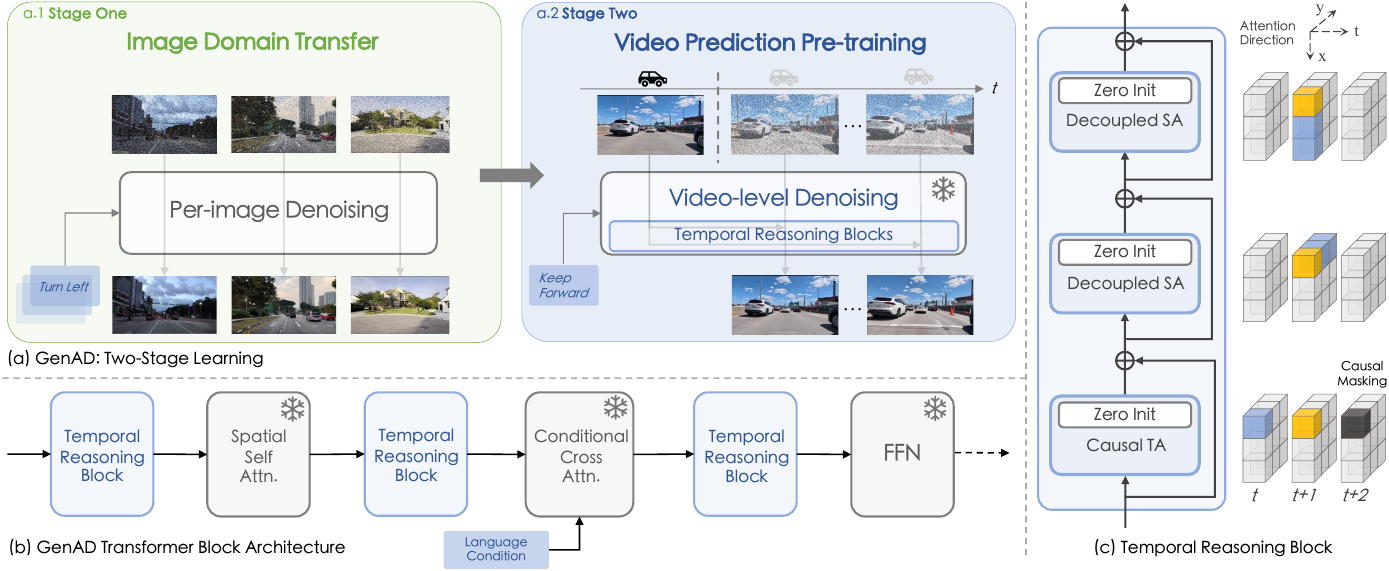}
    \vspace{-20pt}
    \caption{\textbf{Framework of \modelname}. (\textbf{a}) The two-stage learning for \modelname is composed of transferring the image domain of an image diffusion model to the driving field (a.1 Stage one), and video prediction pre-training for modeling the temporal dependency of videos (a.2 Stage two). 
    (\textbf{b}) One transformer block in \modelname for the second stage training has interleaved temporal reasoning blocks before each frozen layer to align spatiotemporal features. 
    (\textbf{c}) The proposed Temporal Reasoning Block includes one causal temporal attention (TA) and two decoupled spatial attention (SA) layers to extract features in different axes. A \textcolor{YellowOrange}{query grid} attends to itself as well as \textcolor{LightCobaltBlue}{blue grids} while the \textcolor{BlackOlive}{dark gray} grid is masked out in causal attention. 
    `{\cmtt{Zero init}}' is appended 
    at the end of each attention block
    to stabilize training.
    }
    \label{fig:pipeline}
\end{figure*}

\section{\modelname Framework}
\label{sec:genad_framework}
In this section, we introduce the training and design of the \modelname model. 
As shown in \cref{fig:pipeline}, \modelname is trained in two stages, \ie, image domain transferring and video prediction pre-training. 
The first stage adapts the general text-to-image model to the driving domain (\cref{subsec: stage1_learning}).
The second stage lifts the text-to-image model to a video prediction model with our proposed temporal reasoning block and modified training schemes (\cref{subsec: stage2_learning}).
In \cref{subsec: extensions}, we explore how the predictive model can be extended to action-conditioned prediction and planning.

\subsection{Image Domain Transfer}
\label{subsec: stage1_learning}
On-board cameras
capture a large field of views with abundant visual contents, including the road, background buildings, surrounding vehicles, \etc, which require strong and robust generation capability to produce continuous and realistic driving scenarios. To facilitate the learning process, we start with independent image generation in the first stage. Concretely, we initialize our model with \sdxl~\cite{Podell2023SDXL}, which is a large-scale latent diffusion model (LDM) for text-to-image generation, to leverage its ability to synthesize high-quality images with plenty of visual details. It is implemented as a denoising \unet $\rvf_\theta$ with several stacked convolution and attention blocks, which learns to synthesize images by denoising the noisy latents~\cite{Rombach2022LDM}.
Specifically, given a noisy input latent $\rvx_{t}$ corrupted by the forward diffusion process, it is trained to predict the added noise $\rvepsilon$ of $\rvx_{t}$ via the following objective:
\begin{equation}
\mathcal{L}_\text{img} \coloneqq \mathbb{E}_{\rvx, \rvepsilon \sim \mathcal{N}(0, 1), \rvc, t}\Big[ \Vert \rvepsilon - \rvf_\theta(\rvx_{t};\rvc,t) \Vert_{2}^{2}\Big] ,
\label{eq:ldmloss}
\end{equation}
where $\rvx$ and $\rvx_{t}$ are the clean and noisy latent, respectively, $t$ denotes the timestep for different noise scales, and $\rvc$ is the text condition that guides the denoising process, 
which is a concatenation of context and command.
For training efficiency, the learning process takes place in a compressed latent space~\cite{Esser2021VQGAN, Rombach2022LDM, Podell2023SDXL} instead of pixel space.
During sampling, the model generates images from standard \gauss noise by denoising the last-step predictions iteratively.

However, the original \sdxl is trained on data in the general domain, such as portraits and artistic paintings, which are not concerned with autonomy systems. 
To adapt the model to synthesize images for driving, we fine-tune it on text-to-image generation using image-text pairs in \datasuitename with the same objective as \cref{eq:ldmloss}. Following the original training of \sdxl, all parameters $\theta$ of the \unet are fine-tuned at this stage, whereas the CLIP text encoders~\cite{Radford2021CLIP} and the autoencoder~\cite{Esser2021VQGAN} remain frozen.

\subsection{Video Prediction Pre-training}
\label{subsec: stage2_learning}

In the second stage, with a few frames of a consecutive video as past observations, \modelname is trained to reason about all visual observations and predict several future frames in plausible ways. Similar to the first stage stage, the prediction process can also be guided by text conditions.
However, predicting the highly dynamic driving world temporally is challenging due to two fundamental barriers. 
\begin{enumerate}[leftmargin=*,itemsep=0pt,topsep=1.3pt]
    \item \textit{Causal Reasoning}: To predict plausible futures following the temporal causality of the driving world, the model needs to comprehend the intentions of all other agents together with the ego vehicle, and understand underlying traffic rules, \eg, how the traffic will change with the transition of traffic lights.
    \item \textit{Drastic View Shift}: 
    Contrary to typical video generation benchmarks which mainly have a static background with slow motion of centered objects, the view of driving changes drastically over time.
    Each pixel in every frame may move to a distant location in the next frame.
\end{enumerate} 
We propose temporal reasoning blocks to address these problems. As illustrated in \cref{fig:pipeline}(c), each block is composed of three successive attention layers, \ie, the causal temporal attention layer and two decoupled spatial attention layers, which are tailored for the causal reasoning and modeling large shifts in the driving scenes, respectively.

\smallskip
\noindent\textbf{Causal Temporal Attention.} 
Since the model after the stage-one training can only process each frame independently, we leverage temporal attention to exchange information among different video frames. 
The attention takes place in the time axis and models the temporal dependency of each grid-wise feature.
However, directly adapting bidirectional temporal attention here as~\cite{Ho2022VideoDiffusionModels,Blattmann2023VideoLDM,Zhang2023Show-1,wang2023LaVie} can hardly acquire the ability of causal reasoning, since the predictions will be inevitably dependent on the subsequent frames instead of past conditions.
Therefore, we restrict the attention direction by adding a causal attention mask, as shown in the last row of \cref{fig:pipeline}(c), to encourage the model to fully exploit knowledge from past observations and faithfully reason about the future as if in real-world driving. We empirically found that the causality constraint greatly regularizes the predicted frames to be coherent with past frames.
Following common practice, we also add temporal bias implemented as relative position embeddings on the time axis~\cite{Shaw2018relativepos} to distinguish different frames of a sequence for temporal attention.

\smallskip
\noindent\textbf{Decoupled Spatial Attention.} 
As driving videos feature fast perspective changes, 
features in a specific grid could vary greatly in different timesteps 
and are hard to correlate and learn by temporal attention, which suffers from a limited receptive field. In light of this, we introduce spatial attention to propagate each grid feature in spatial axes to aid in gathering information for temporal attention. We implement a decoupled variant of self-attention 
for its efficiency with linear computational complexity, compared to quadratic full self-attention. As shown in \cref{fig:pipeline}(c), the two 
decoupled attention layers propagate features in horizontal and vertical axes, respectively.

\smallskip
\noindent\textbf{Deep Interaction.} Intuitively, the spatial blocks fine-tuned in stage one refine features of each frame independently towards photorealism, whereas the temporal blocks introduced in stage two align features of all video frames towards coherency and consistency. To further boost the spatiotemporal feature interaction, we interleave the proposed temporal reasoning blocks with the original Transformer blocks in \sdxl, \ie, spatial attention, cross attention, and feed-forward network, as shown in \cref{fig:pipeline}(b).

\smallskip
\noindent\textbf{Zero Initialization.}
Similar to the previous practices~\cite{Zhang2023ControlNet, Alayrac2022Flamingo}, for each block that is newly introduced in stage two, we initialized all parameters of its final layer as zero. This avoids disrupting the prior knowledge of the well-trained image generation model in the beginning and stabilizes the training process.

\smallskip
\noindent\textbf{Training.} 
\modelname is trained to predict the future by jointly denoising from the noisy latents with the guidance of past frames and text conditions.
We first project $T$ consecutive frames of a video clip into a batch of latents $\rvv\!=\!\{\rvv^m, \rvv^n\}$, where the leading $m$ frame latents $\rvv^m$ are clean, representing historical observations, and other $n\!=\!T\!-\!m$ frame latents $\rvv^n$
indicate the future to be predicted.
$\rvv^n$ are then corrupted to $\rvv^n_t$ by the forward diffusion process, where $t$ indexes a randomly sampled noise scale. The model is trained to predict the noise of $\rvv^n_t$ 
conditioned on observations $\rvv^m$ and text $\rvc$.
The learning objective of the video prediction model is formulated as follows:
\begin{equation}
\mathcal{L}_\text{vid} \coloneqq \mathbb{E}_{\rvv, \rvepsilon \sim \mathcal{N}(0, 1), \rvc, t}\Big[ \Vert \rvepsilon - \rvf_{\theta, \phi}(\rvv^n_{t};\rvv^m, \rvc,t) \Vert_{2}^{2}\Big] ,
\label{eq:video-ldmloss}
\end{equation}
where $\theta$ denotes the inherited stage-one model and $\phi$ represents the newly inserted temporal reasoning blocks. Following \cite{Blattmann2023VideoLDM}, we freeze $\theta$ and only train the temporal reasoning blocks to avoid perturbing the generation ability of the image generation model and focus on learning temporal dependencies in videos.
Note that only the outputs from the corrupted frames $\rvv^n_t$ contribute to the training loss while those from condition frames $\rvv^m$ are ignored. Our training recipe is also readily applicable to video interpolation with minor modifications, \ie, switching the indices of condition frames.
\subsection{Extensions}
\label{subsec: extensions}
Relying on the well-trained video prediction capability in driving scenarios, we further exploit the potential of the pre-trained model in action-controlled prediction and planning, which are important for real-world driving systems. Here, we explore the downstream tasks on \nusdata~\cite{Caesar2019nuScenes} which provides recorded poses.

\smallskip
\noindent\textbf{Action-conditioned Prediction.}
To make our predictive model controllable with exact ego actions and act as a simulator~\cite{Kim2021DriveGAN}, we fine-tune the model with the paired future trajectory as an additional condition. Specifically, we map the raw trajectory to a high-dimensional feature with Fourier embeddings~\cite{tancik2020Fourier}. After further projection by a linear layer, it is added to the original conditions. Thus, the ego actions are injected into the network through the conditional cross-attention layer in \cref{fig:pipeline}(b).

\smallskip
\noindent\textbf{Planning.}
By learning to predict the future, \modelname acquires strong representations of complex driving scenes, which can be further exploited for planning.
Specifically, we extract spatiotemporal features of two historical frames through the UNet encoder of the \textit{frozen} \modelname, which is nearly half the size of the entire model,
and feed them to a multi-layer perceptron (MLP) to predict future waypoints. 
With the frozen \modelname encoder and a learnable MLP layer, 
the training process of our planner can be sped up by 3400 times compared to an end-to-end planning model UniAD~\cite{Hu2023UniAD}, validating the effectiveness of the learned spatiotemporal feature of \modelname.

\begin{figure*}[t!]
    \centering
    \includegraphics[width=\textwidth]{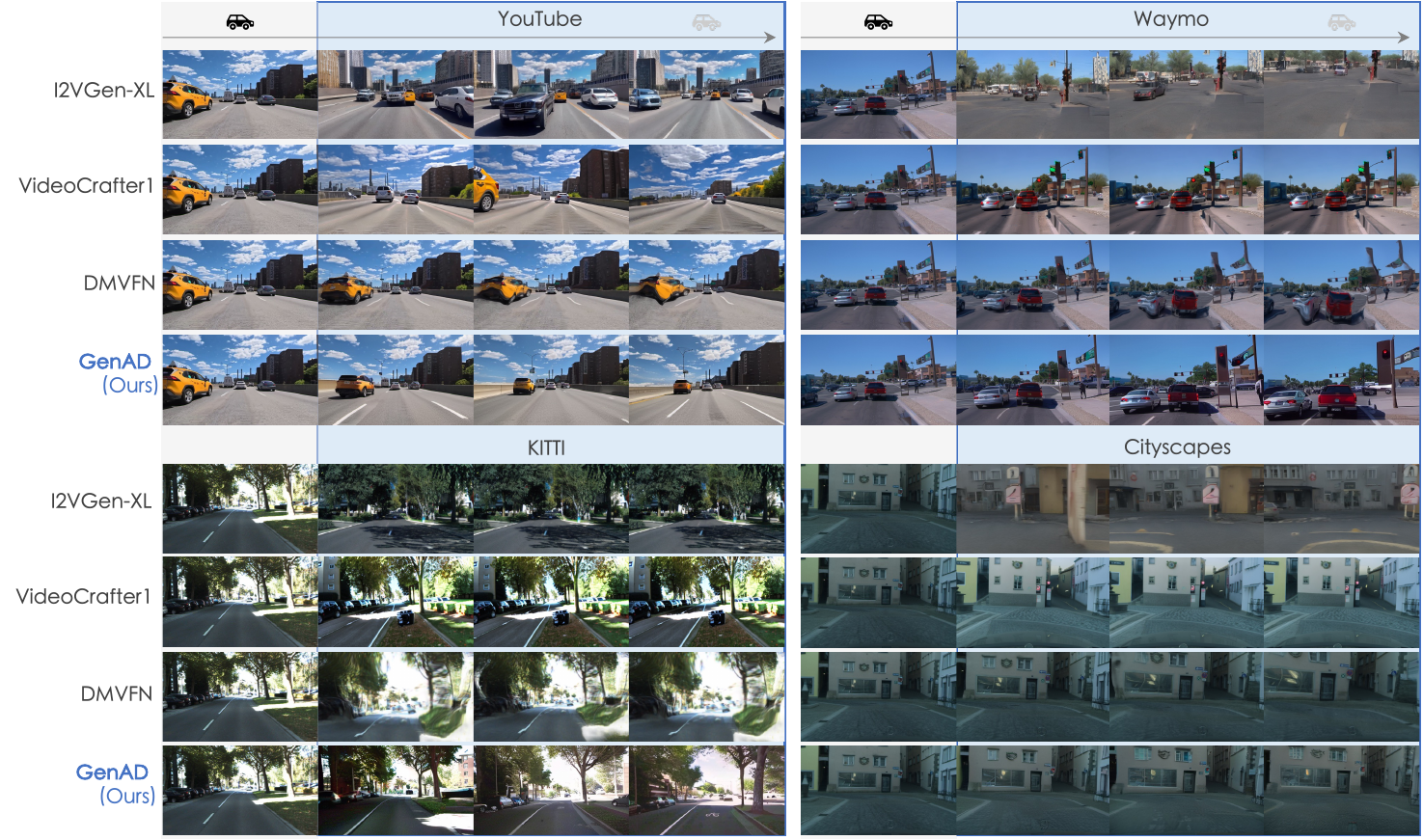}
    \vspace{-20pt}
    \caption{\textbf{Task on zero-shot video prediction for unseen scenarios}. We show the generation results (in blue boxes) of different models given the same starting frames. \modelname makes more robust, realistic, and reasonable future predictions on unseen datasets (scenarios). More comparisons (\cref{fig:zero-shot-public}) and visualizations (\cref{fig:zero-shot-youtube}) are shown in Appendix.
    }
    \vspace{-.3cm}
    \label{fig:compare-videogen}
\end{figure*}

\section{Experiments}
\label{sec:exp}

\subsection{Setup and Protocols}
\modelname is learned in two stages on \datasuitename but with different learning objectives (in \cref{sec:genad_framework}) and input formats.
In stage one, the model takes input (image, text) pairs and is trained on text-to-image generation. We broadcast the command annotation, which is labeled for each 4s video sequence, to all frames included. The model is trained for 300K iterations on 32 NVIDIA Tesla A100 GPUs with a total batch size of 256.
In the second stage, \modelname is trained to jointly denoise future latents conditioned on past latents and texts. Its inputs are (video clip, text) pairs where each video clip is 4s at 2Hz. 
The current version of \modelname
is trained on 64 GPUs for 112.5K iterations with a total batch size of 64.
The input frames are resized to $256\!\times\!448$ for training in both stages, and the text condition $\rvc$ is dropped at a probability of $p\!=\!0.1$ to enable classifier-free guidance~\cite{Ho2021classifierfree} in sampling, which is commonly used in diffusion models to improve sample quality.
More training and sampling details are in \cref{supp:implementation}.
\subsection{Results of Video Prediction Pre-training}

\noindent\textbf{Comparison to Recent Video Generation Approaches.} We compare \modelname to recent advances on an unseen set with geofencing from \youtubesplitename, Waymo~\cite{Sun2019WOD}, KITTI~\cite{Geiger2013KITTI}, and Cityscapes~\cite{Cordts2016Cityscapes} in a \textit{zero-shot} generation manner. \cref{fig:compare-videogen} depicts the qualitative results. 
Image-to-video models I2VGen-XL~\cite{Zhang2023i2vgen} and VideoCrafter1~\cite{Chen2023VideoCrafter1} can not strictly follow the given frames to make predictions, yielding poor consistency between the predicted frames and past frames. 
The video prediction model DMVFN~\cite{Hu2023DMVFN} that is trained on Cityscapes suffers from the unfavorable shape 
distortions in its predictions, especially on the three unseen datasets. In contrast, \modelname exhibits remarkable zero-shot generalization ability 
and 
visual quality
although \textit{none} of these sets are included in the training.

\begin{table}[t!]
    \footnotesize
    \centering
    \vspace{.12cm}
    \begin{tabular}{l|c|c|cc}
        \toprule
        \multirow{2}{*}{Method} & \multirow{2}{*}{\makecell{Training\\Dataset}} & \multirow{2}{*}{Pred.} & \multicolumn{2}{c}{\nusdata} \\
        &  &  & FID ($\downarrow$) & FVD ($\downarrow$) \\
        \midrule
        DriveGAN~\cite{Kim2021DriveGAN} & \multirow{3}{*}{\nusdata} & \cmark & 73.4 & 502 \\ 
        DriveDreamer$^*$~\cite{Wang2023DriveDreamer} & & \cmark & 52.6 & 452 \\
        DrivingDiffuion$^*$~\cite{Li2023DrivingDiffusion} & & \xmark & 15.8 & 332 \\
        \midrule
        \modelname-nus (Ours) & \nusdata &  \cmark & \textbf{15.4} & 244 \\
        \modelname (Ours) & \datasuitename &  \cmark & \textbf{15.4} & \textbf{184} \\
        \bottomrule
    \end{tabular}
    \caption{\textbf{Video generation quality compared to state-of-the-arts trained on \nusdata}. ``Pred.": evaluation by future prediction. $^*$: requiring 3D layout inputs.
    }
    \label{tab:sota_nus}
    \vspace{-1em}
\end{table}

\begin{figure}[t!]
    \centering
    \includegraphics[width=\linewidth]{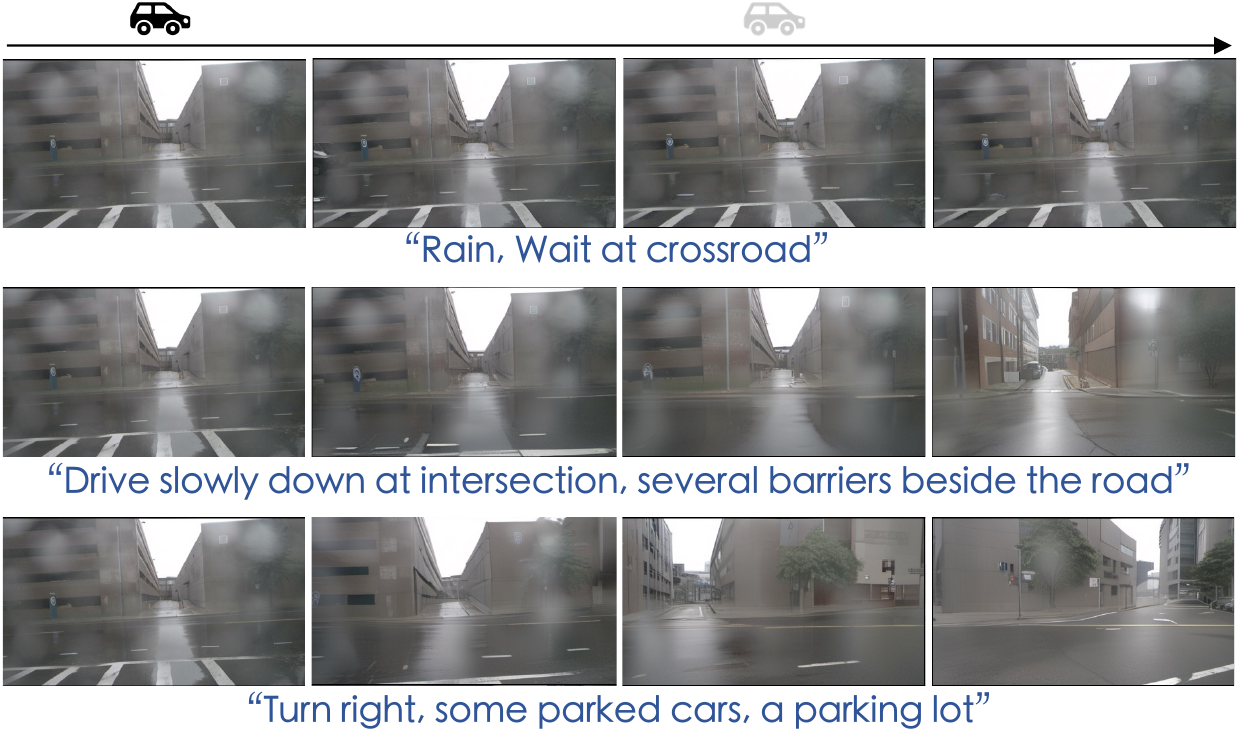}
    \caption{\textbf{Task on langauge-conditioned prediction}.
    Given two frames of a rainy scenario in the intersection and three high-level text conditions, \modelname simulates reasonable futures accordingly.
    }
    \vspace{-.4cm}
    \label{fig:language_condition}
\end{figure}

\smallskip
\noindent\textbf{Comparison to \nusdata Experts.}
We also compare \modelname with the most recent available driving video generation models which are exclusively trained for \nusdata. \cref{tab:sota_nus} shows that \modelname surpasses all previous methods in both image fidelity (FID) and video coherence (FVD). Specifically, \modelname significantly reduces FVD by \textbf{44.5\%} compared to DrivingDiffusion~\cite{Li2023DrivingDiffusion}, without taking 3D future layouts as additional inputs. For fair comparisons, we train a model variant (\modelname-nus) on \nusdata dataset only. We find that although \modelname-nus performs on par with \modelname on \nusdata, it struggles to generalize to unseen datasets like Waymo, where the generation degrades to the \nusdata visual pattern.
In contrast, \modelname trained on \datasuitename exhibits strong generalization ability across datasets as shown in \cref{fig:compare-videogen}.

We provide language-conditioned prediction samples on \nusdata in \cref{fig:language_condition}, where \modelname simulates various futures from the same start following different textual instructions. The impressive generation quality is exhibited in the intricate details of the environment, and the natural transition of ego motion. 

\begin{figure}[t!]
    \centering
    \includegraphics[width=\linewidth]{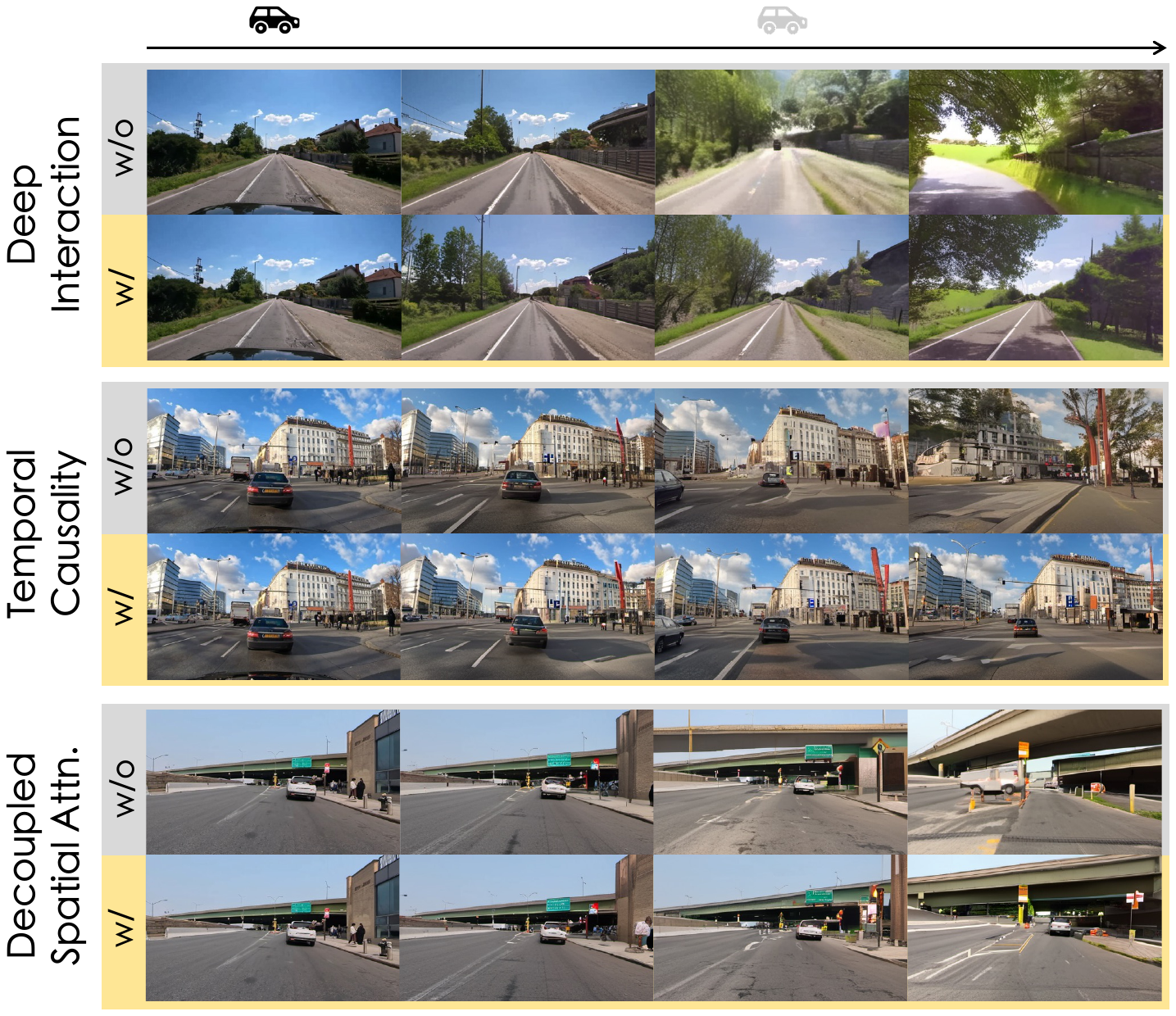}
    \vspace{-16pt}
    \caption{\textbf{Case study for model designs}. All components help alleviate artifacts and improve the consistency of future predictions.}
    \label{fig:ablation}
\end{figure}

\smallskip
\noindent\textbf{Ablation Study.}
We perform ablations by training each variant on a subset of \datasuitename for 75K steps. Starting from the baseline with plain temporal attentions~\cite{Blattmann2023VideoLDM,Ho2022VideoDiffusionModels}, we gradually introduce our proposed components. Notably, by interleaving the temporal blocks with the spatial blocks,
the FVD significantly improves (-17\%) due to more sufficient spatiotemporal interactions. Both temporal causality and decoupled spatial attention contribute to better CLIPSIM, improving the temporal consistency between future predictions and the condition frames. 
To be clear, 
the slight increase in FID and FVD, shown in fourth and third rows of~\cref{tab:ablation} respectively, does not faithfully reflect a decline in generation quality as discussed in~\cite{Blattmann2023VideoLDM, Brooks2022LongVideoGAN, Podell2023SDXL}. The effectiveness of each design is shown in \cref{fig:ablation}.

\begin{table}[t!]
    \footnotesize
    \centering
    \begin{tabular}{l|ccc}
        \toprule
        \multirow{2}{*}{Method} & \multicolumn{3}{c}{YouTube} \\
        & FID ($\downarrow$) & FVD ($\downarrow$) & CLIPSIM ($\uparrow$) \\
        \midrule
        Baseline & 18.32 & 244.44 & 0.8405 \\
        $+$ Deep Interaction & 17.96 & 201.69 & 0.8409 \\
        $+$ Temporal Causality & \textbf{16.54} & 207.45 & 0.8550 \\
        $+$ Decoupled Spatial Attn. & 17.67 & \textbf{189.54} & \textbf{0.8652} \\
        \bottomrule
    \end{tabular}
    \vspace{-6pt}
    \caption{\textbf{Ablation on model designs in \modelname}.
    All proposed designs contribute to the final performance.
    }
    \label{tab:ablation}
\end{table}

\subsection{Results of Extensions}
\begin{figure}[t!]
    \centering
    \includegraphics[width=\linewidth]{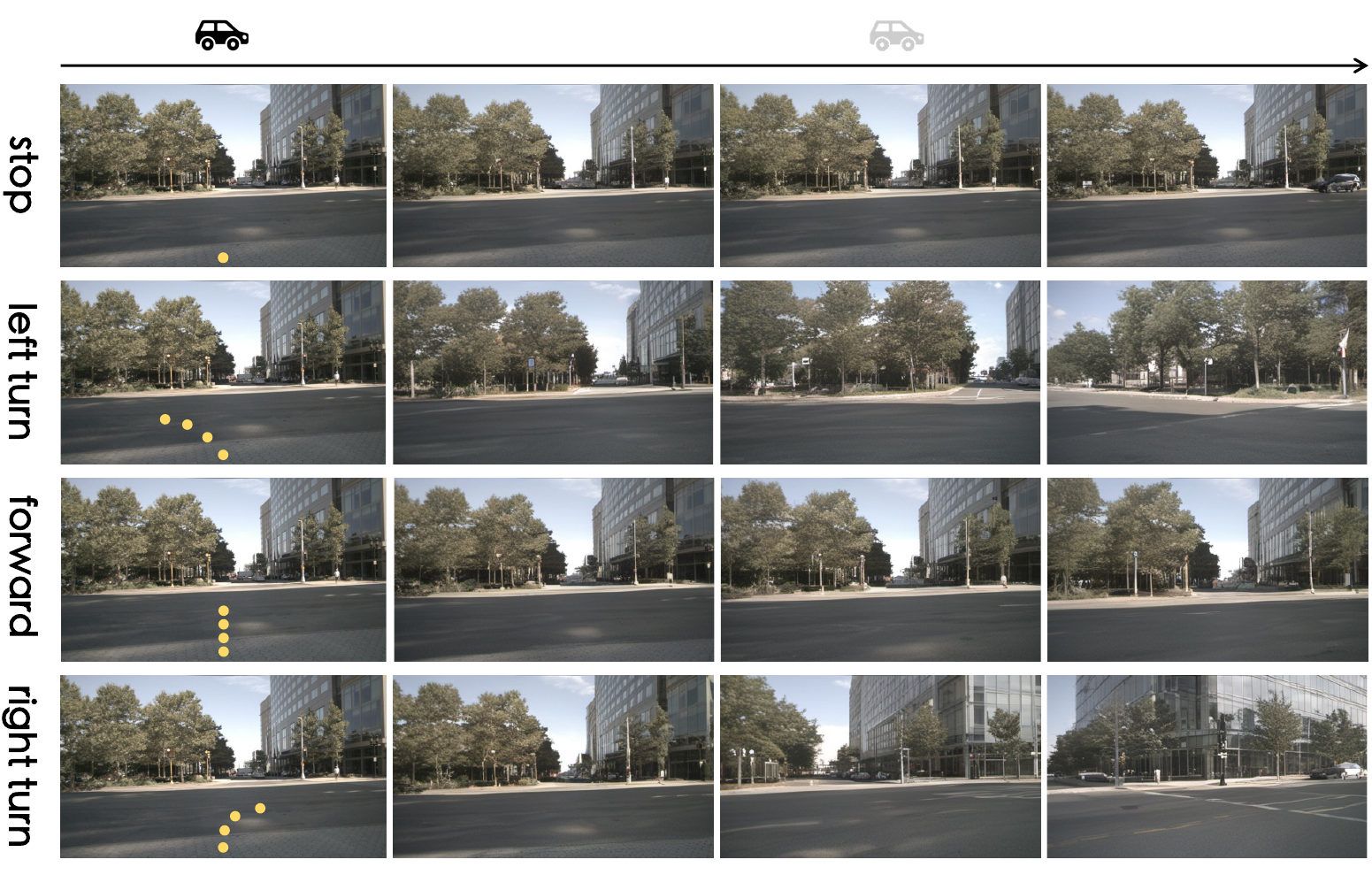}
    \vspace{-16pt}
    \caption{\textbf{Task on action-conditioned prediction (simulation)}. 
    Given the same starting frames and different future trajectories (shown in yellow dots in the first column), \modelname-act can simulate diverse futures following different ego intentions. More visualizations are in Appendix~\cref{fig:action}.
    }
    \label{fig:action-vis}
\end{figure}

\noindent\textbf{Action-conditioned Prediction.}
We further showcase
the performance of the action-conditioned model fine-tuned on \nusdata, \modelname-act, in \cref{fig:action-vis} and \cref{tab:res_action_condition}.
Given two starting frames and a trajectory $\mathbf{w}$ composed of 6 future waypoints, \modelname-act imagines 6 future frames following the trajectory sequence.
To evaluate the consistency between the input trajectory $\mathbf{w}$ and predicted frames, we establish an inverse dynamics model (IDM) on \nusdata as the evaluator, which projects a video sequence into a corresponding ego trajectory.
We leverage the IDM to translate predicted frames into the trajectory $\hat{\mathbf{w}}$, and calculate the L2 distance between $\mathbf{w}$ and $\hat{\mathbf{w}}$ as the Action Prediction Error.
Specifically, \modelname-act substantially reduces the Action Prediction Error by 20.4\% compared to \modelname with text condition, allowing for more accurate future simulations.

\begin{table}[t!]
    \footnotesize
    \centering
    \begin{tabular}{l|c|c}
        \toprule
        \multirow{2}{*}{Method} & 
        \multirow{2}{*}{Condition} & 
        \nusdata \\
        & & Action Prediction Error ($\downarrow$) \\
        \midrule
        \textcolor{gray}{Ground truth} & \textcolor{gray}{-} & \textcolor{gray}{0.90} \\
        \modelname & text & 2.54 \\
        \modelname-act & text + traj. & \textbf{2.02} \\ 
        \bottomrule
    \end{tabular}
    \vspace{-6pt}
    \caption{\textbf{Task on action-conditioned prediction}. 
    Compared to \modelname with text conditions only, 
    \modelname-act enables more precise future predictions that follow the action condition.  
    }
    \label{tab:res_action_condition}
    \vspace{-.3cm}
\end{table}

\begin{table}[t!]
    \footnotesize
    \centering
    \begin{tabular}{l|c|ccc}
        \toprule
        \multirow{2}{*}{Method} & \multirow{2}{*}{\makecell{\# Trainable\\Params.}} &
        \multicolumn{2}{c}{\nusdata} \\
        & & ADE ($\downarrow$) & FDE ($\downarrow$) \\
        \midrule
        ST-P3$^*$~\cite{hu2022stp3} & 10.9M & 2.65 & 3.73 \\
        UniAD$^*$~\cite{Hu2023UniAD} & 58.8M & 1.03 & 1.65 \\
        \midrule
        \modelname (Ours) & 0.8M & 1.23 & 2.31\\
        \bottomrule
    \end{tabular}
    \vspace{-6pt}
    \caption{\textbf{Task on open-loop planning}. A lightweight MLP with \textit{frozen} \modelname gets competitive planning results with 73$\times$ fewer trainable parameters and front-view image alone.
    $^*$: multi-view inputs.
    Evaluation protocols are aligned with UniAD~\cite{Hu2023UniAD}.
    }
    \label{tab:res_plan}
    \vspace{-.3cm}
\end{table}

\smallskip
\noindent\textbf{Planning Results.}
\cref{tab:res_plan} depicts the planning results on \nusdata
where ground truth poses for the ego vehicle are available.
By freezing \modelname encoder and only optimizing an additional MLP on top of it, 
the model can effectively learn to plan.
Notably, by pre-extracting image features through the UNet encoder of \modelname, the entire 
learning process for planning adaptation takes only 10 minutes on a single NVIDIA Tesla 
V100 device, which is 3400 times more efficient than the training of the UniAD planner~\cite{Hu2023UniAD}.

\section{Limitations and Discussion}
\label{sec:discussion}

We study the system-level development of \modelname, a large-scale generalized video predictive model for autonomous driving.
We also validate the adaptation of the learned representation of \modelname to driving tasks, \ie, learning a ``world model'' and motion planning.
Although we obtain improved generalization to open domains,
the increased model capacity poses challenges in both training efficiency and real-time deployment.
We envision the unified video prediction task will serve as a scalable objective for future research on representation learning and policy learning. Another interesting direction involves distilling the encoded knowledge for a wider range of downstream tasks~\cite{Li2023DreamTeacher}.

\section*{Acknowledgements}
OpenDriveLab is the autonomous driving team affiliated with Shanghai AI Lab.
This work was supported by National Key R\&D Program of China (2022ZD0160104), NSFC (62206172), Shanghai Committee of Science and Technology (23YF1462000), and China Postdoctoral Science Foundation (2023M741848). This work was also partially supported by the BMBF (Tübingen AI Center, FKZ: 01IS18039A), the DFG (SFB 1233, TP 17, project number: 276693517), and by EXC (number 2064/1 – project number 390727645).
We thank team members from OpenDriveLab for valuable feedback along the project.

{
    \small
    \bibliographystyle{ieeenat_fullname}
    \bibliography{bibliography_short, bibliography}
}

\appendix
\onecolumn

\newcommand\DoToC{%
  \startcontents
  {
      \hypersetup{linkcolor=black}
      \printcontents{}{1}{\noindent\textbf{\Large Appendix}\vskip5pt}
  }
}
\maketitle
\DoToC

\clearpage \newpage

\section{Discussions}
\label{supp:discussion}

To assist a better understanding of our work, we supplement discussions on 
intuitive questions that one may raise.

\bigskip
\noindent\textbf{Q1.} \textit{Why do you propose the training target of the predictive model as videos?
}
\smallskip

Video is a particularly universal and scalable target given a wealth of uncalibrated driving videos. Different from BEV representations~\cite{Hu2022MILE,Gao2022SEM2} that require camera extrinsic parameters and point clouds~\cite{khurana2023point,weng2021SPF2} that are restricted by different LiDAR configurations, video prediction can be performed in a pose-agnostic manner. This characteristic offers significant advantages in scalability to more diverse data sources, which are key for the generalization ability of the learned model.

\bigskip
\noindent\textbf{Q2.} \textit{Why do you predict multiple frames simultaneously with historical frames as input? How about alternatively using an auto-regressive design, i.e., predicting future frames one by one?
}
\smallskip

Indeed, auto-regressive prediction can further stabilize the prediction process by leveraging conditional dependencies on previously generated frames, thereby enhancing consistency. Nevertheless, we still choose to employ a joint denoising procedure for two primary reasons.
To start, diffusion models are typically computationally expensive, and our model is no exception. For videos comprised of multiple frames, predicting them auto-regressively would multiply their computational intensity, making it inefficient for implementation and deployment.

Moreover, conducting auto-regressive predictions makes it challenging to effectively apply conditions that require significant changes. Consider the scenario of a driver making a turn, which typically takes several seconds and involves a long sequence of frames. If the prediction duration is too short, the model may struggle to follow the given instructions, as it is impossible to achieve substantial changes within a single frame. Instead, it might simply continue the tendency of previously determined frames and completely disregard the provided instructions. Therefore, joint prediction also allows us to effectively apply complex controls and facilitate more coherent action generation.

\bigskip
\noindent\textbf{Q3.} \textit{What is the criterion to prove good generalization ability of your model? How much data do we need to guarantee generalization?
}
\smallskip

Currently, it is hard to define a specific criterion to assess the generalization ability of our predictive models for the reason that the quality judgment is subjective~\cite{liu2023EvalCrafter} and it is impossible to find an aligned method that is available to compare. However, through our exhaustive exploitation of public data, we have discovered that increasing the scale of the data is advantageous for zero-shot generation on existing datasets. It is also important to note that our method is easily scalable, offering opportunities to continuously enhance its generalization ability by leveraging vast amounts of unlabeled data.

\bigskip
\noindent\textbf{Q4.} \textit{Why not evaluate models using typical video prediction metrics? What are the appropriate metrics to evaluate the performance of the driving video prediction model with multiple conditions?}
\smallskip

Common practices in the task of video prediction use Structural Similarity Index Measure (SSIM)~\cite{Wang2003SSIM} and a perceptual metric LPIPS~\cite{zhang2018LPIPS} for quantitative evaluation. These two metrics calculate frame-wise similarities between predicted frames and corresponding ground truth frames. They are designed to assess the model's ability to \emph{exactly} follow the recorded events.
Consequently, models optimized for these metrics tend to copy certain patterns and could overfit the small datasets adopted, thereby restricting their potential for diverse future generations. This limitation is particularly problematic for predictive models in driving scenarios, where multiple futures may occur and proactive preparation is essential for each of these cases. 

We sought to use the distribution-based metrics, including FVD~\cite{Unterthiner2018FVD} and CLIPSIM which are widely adopted by diffusion-based generation approaches~\cite{Zhang2023Show-1,Blattmann2023VideoLDM,wang2023LaVie}. However, for the image-to-video generation models~\cite{Chen2023VideoCrafter1,Zhang2023i2vgen} in our comparison, they do not directly compose the input image as any specific video frames they generate, mainly preserving semantics and contents from input images. Thus, it becomes challenging to align the comparison settings with ours for metrics like FVD, which measures the distribution distance of consecutive frames, or CLIPSIM, which can be used to evaluate the semantic similarity between the conditional frame and generated frames. Moreover, these metrics are not perfect. For instance, FVD could be blind to unrealistic repetition and prefer small-scale motion, as discussed in \cite{Brooks2022LongVideoGAN,Blattmann2023VideoLDM}.

In short, from the existing metrics, it is hard to quantitatively evaluate the prediction abilities of a generalized model for real-world driving, which encompasses multi-modal conditions and requires temporal consistency. There still needs to be effort made to design an appropriate metric that can effectively evaluate such models.

\bigskip
\noindent\textbf{Q5. Broader impact.} \textit{What are potential applications and future directions with the provided large-scale \datasuitename data and the \modelname model, for both academia and industry?}
\smallskip

To the best of our knowledge, \datasuitename is the largest available data corpus that we can collect from public sources. It significantly enhances the quantity and diversity of driving video footage in multiple dimensions, providing the research community with a massive high-quality resource for exploring open avenues in autonomous driving. In addition to video prediction, we hope our dataset can also benefit the community to enable broader applications~\cite{radosavovic2023real,wang2023all,Dai2023UniPi,Yang2023UniSimRobo,Shah2023ViNT}.

In this work, we have demonstrated that the strong representation of \modelname can be beneficial for planning. Similarly, it is also promising to adapt it to a broader range of downstream tasks such as perception~\cite{zhao2023VPD}. To improve the flexibility and efficiency for deployment, transferring the knowledge of generative models via distillation~\cite{Li2023DreamTeacher} is also worth investigating. Except for its powerful representations, the prediction futures conditioned on actions also open the opportunities for model-predictive control~\cite{Gupta2023MaskViT,lecun2022PATH} and inverse dynamics model~\cite{Dai2023UniPi,baker2022VPT,Ajay2023HiP} to enable trajectory planning, which are beyond the scope of this paper. Note that our model will be made publicly available to benefit the community and it is flexible to further fine-tune it on in-house data for the industry.

\bigskip
\noindent\textbf{Q6. Limitations.} \textit{What are the issues with current designs, and corresponding preliminary solutions?}
\smallskip

It is known that the captions used for training have a great impact on generation quality~\cite{chen2023PixArt,betker2023DALLE3}. Currently, the context description of \datasuitename is automatically annotated by BLIP-2~\cite{Li2023BLIP2}. However, we empirically find that the generated captions have two main limitations. First, the BLIP-2 captions tend to be short and plain, lacking enough details about the complicated driving scenes and becoming indistinguishable from one another. Second, the alignment between the image and caption still needs to be improved. The BLIP-2 captions are mostly centric on a single object and thus fail to include the majority of important content in the scene. In addition, affected by its fine-tuning samples~\cite{lin2014COCO}, BLIP-2 is unaware of the state of the image observer itself. Hence, it fails to infer ego intentions, which may lead to conflicts with high-level commands. To overcome these limitations, it is promising to utilize more advanced vision-language models that have a more comprehensive understanding and text-rich description of the whole scene~\cite{OpenAI2022GPT4,liu2023LLaVA,gao2023LLaMAAdapterV2}, and have temporal awareness~\cite{li2023Otter}.

We opt for SDXL~\cite{Podell2023SDXL} as our starting point to inherit its merits in high quality of visual details, large capacity of model size, and better rendering abilities of text encoders. On the other hand, we have noticed that SDXL is slow to sample and computationally expensive. Our model does suffer from that as well. However, as a pioneering work exploring how to build a generalized predictive model on internet-scale driving data, the main focus of this work is the generalization ability to diverse unseen driving scenarios instead of computation overhead. Future works may include trying faster sampling methods~\cite{zhao2023UniPC,meng2023distillation,salimans2022progressive} and transferring our general recipe to more efficient diffusion models~\cite{Rombach2022LDM}.

\medskip
While there is not a silver bullet yet we hope that future work takes a deep and grounded look at these discussions, identifying what more downstream applications could be applied - and more importantly, why they work or fail. Our hope is that \modelname serves as a starting point, as the main paper argues, a generalized video pre-trained paradigm that is built on top of the largest available driving videos and excels at a wide spectrum of autonomous driving tasks.

\section{Related Work}
\label{supp:related}

Related work is introduced below due to the limited space in the main paper.

\subsection{Driving Scene Generation}

Over the past few years, scene generation has gained increasing popularity due to its importance for safety-critical domains like autonomous driving. One family of works~\cite{Chen2021GeoSim,Xu2023dDisCoScene,Yang2023UniSimAD, Yang2023Vidar} perform 3D-aware rendering for sensor simulation. In particular, GeoSim~\cite{Chen2021GeoSim} augments the existing images by borrowing objects from other scenes and rendering them at novel poses. UniSim~\cite{Yang2023UniSimAD} creates digital twins of driving logs with manipulable foreground objects to enable close-loop simulation. However, these methods can only manipulate objects from the collected assets, and novel objects cannot be created unless further collected. Recent advancements~\cite{Kim2023NFLDM,Swerdlow2023BEVGen,Yang2023BEVControl,chen2023GeoDiffusion,Gao2023MagicDrive,Li2023DrivingDiffusion,Wang2023DriveDreamer} use diffusion models to synthesize scenes with novel content beyond the collected data. As a dual task of perception, several works simulate realistic sensor data controlled by input layouts such as 2D bounding boxes~\cite{chen2023GeoDiffusion} and bird's-eye view (BEV) segmentation maps~\cite{Swerdlow2023BEVGen,Yang2023BEVControl,Kim2023NFLDM}. More recent works~\cite{Wang2023DriveDreamer,Li2023DrivingDiffusion,Gao2023MagicDrive} choose 3D bounding boxes for better geometry control. These methods also have potential to serve as data engine~\cite{Yang2023UniSimAD,Swerdlow2023BEVGen,chen2023GeoDiffusion,Li2023DrivingDiffusion,Gao2023MagicDrive,Yang2023BEVControl}, \ie, the simulated data can be further adopted as augmented samples to boost the performance of existing perception models. However, their control abilities are acquired from manually annotated datasets, preventing them from scaling to more unlabelled data and increasing both diversity and generalization.

Besides layout-controlled sensor simulation, another thread of progresses~\cite{Hu2022MILE,Gao2022SEM2,Kim2021DriveGAN,Wang2023DriveDreamer,Hu2023GAIA} focuses on simulating the temporal dynamics of the driving scenarios. Specifically, MILE~\cite{Hu2022MILE} firstly introduces a model of the world incorporating the BEV representation. By imagining the world within the designed space, the world dynamics can be implicitly encoded and the behaviors of vehicles can be interpretably decoded. This opens the opportunity for executing planning policies without having access to real observations. Differently, inspired by the advances in video generation, DriveDreamer~\cite{Wang2023DriveDreamer} and GAIA-1~\cite{Hu2023GAIA} propose to build a realistic world model in the form of video frames. Particularly, GAIA-1 is scaled up to about 10B model parameters on 4700 hours of in-house videos, showing highly appealing results. However, the diversity of their generation is still limited by the datasets they adopt. To be specific, the nuScenes~\cite{Caesar2019nuScenes} used by DriveDreamer is collected in Singapore and Boston, while GAIA-1’s driving logs are recorded within London. Both of them use fixed or similar camera settings. The distribution of their data sources limits their generalization abilities to unseen scenarios, different camera poses, and other settings. Moreover, how to utilize the learned knowledge for downstream applications, \eg, planning, is still rarely mentioned and explored.

\subsection{Video Generation and Prediction}

Video generation and prediction are effective ways to model the real world. Several practices~\cite{Gupta2023MaskViT,Yu2023MAGVIT,Hu2023DMVFN} have been made to synthesize future driving videos. With the renaissance of diffusion models~\cite{Ho2020DDPM,song2020DDIM}, recent progresses~\cite{Saharia2022imagen,Rombach2022LDM,nichol2021GLIDE,ramesh2022unCLIP} have demonstrated that diffusion models show a great advantage over other generative methods~\cite{goodfellow2014GAN,kingma2013VAE,rezende2015Flow} in both fidelity and diversity. These advantages have also been extended to the temporal domain by numerous works in video generation~\cite{Blattmann2023VideoLDM,He2022LVDM,wang2023LaVie,li2023VideoGen,gu2023VidRD}. Among them, many works~\cite{harvey2022FDM,Voleti2022MCVD,lu2023VDT,Blattmann2023VideoLDM} include public driving datasets~\cite{Geiger2013KITTI,Cordts2016Cityscapes,Caesar2019nuScenes} as touchstones for their evaluation. However, none of these methods have proposed effective designs that are specialized for driving scenarios, which are known to be more complex and challenging~\cite{Gupta2023MaskViT} as we discussed in the main paper. In addition, due to their exclusive training strategy, the model capability is greatly limited by each small and simple dataset~\cite{Geiger2013KITTI,Cordts2016Cityscapes,Caesar2019nuScenes}, hindering the generalization ability to diverse driving scenes in the real world. In contrast, we explore the first practice of building a generalized prediction model via training on large-scale driving videos in a joint manner.

\subsection{Learning from Web Driving Videos}
\label{sec:related-web-driving-videos}

Learning the general capabilities from large-scale data has been well studied in the field of both vision and language~\cite{Schuhmann2022LAION,Brown2020GPT3,Radford2021CLIP}. It is also promising to exploit the internet-scale videos for autonomous driving. However, due to the unlabeled nature of the web data, there exist great challenges and there are only a few methods that leverage this idea to driving tasks for different purposes. SelfD~\cite{Zhang2022SelfD} learns driving policies via semi-supervised learning on YouTube videos. The policy network is pre-trained with pseudo trajectories and then transferred to the target datasets via fine-tuning. Instead of directly pre-training the policy, ACO~\cite{Zhang2022ACO} introduces an action contrastive learning method to obtain action-related representations for downstream tasks. However, both SelfD and ACO rely on pseudo-labeling of trajectories or actions on vast amounts of driving videos. This could be highly sensitive to domain changes, thus compromising their reliability. More recently, PPGeo~\cite{Wu2023PPGeo} proposes a fully self-supervised learning pipeline to learn a motion-aware encoder through geometric reconstruction. The encoder can be further fine-tuned to benefit downstream tasks. However, their pipeline requires separating each component into different training stages. 
Instead,
our method directly conducts self-supervised learning via future prediction, which is more intuitive and flexible. This allows us to easily apply it to such massive and diverse uncalibrated driving videos for the first time. In addition, our predictions generate interpretable visual outputs that implicitly perform the planning process and seamlessly serve as a real-world driving simulator.

\subsection{Video Datasets from the Internet}

Large-scale datasets have been proven to be a core component for generalizable foundation models~\cite{Radford2021CLIP, Schuhmann2022LAION}. For video tasks, collecting data in laboratories or through crowd-sourcing is a common strategy for specific tasks, such as robotics~\cite{Brohan2023RT1} and ego-centric perception~\cite{Grauman2022Ego4D}. However, the collection and annotation process is costly and hard to scale. Therefore, researchers have sought YouTube or similar websites as video sources as they cover diverse topics and environments, and support academic usage licenses. For example, some pioneering works manually annotate YouTube videos for action classifications~\cite{Soomro2012UCF101, Kay2017Kinetics, Qian2022Articulation3D}, action descriptions or captions~\cite{Das2013YouCook, Zhou2018YouCook2}, and hand-object intersections~\cite{Fouhey2018VLOG, Shan2020100DOH}. Recently, researchers have begun to leverage alt-text~\cite{Bain2021WebVid}, automatic speech recognition~\cite{Miech2019HowTo100M, Xue2022HDVILA, Zellers2021MERLOT}, original image captions~\cite{Nagrani2022VideoCC3M}, or paired subtitles~\cite{Sanabria2018How2, Miech2019HowTo100M} to enlarge the annotation scale for video captions. With the development of foundation models, Wang \etal~\cite{Wang2023InternVid} employ image captioning models and language models to generate video captions. These video-text pairs have demonstrated great help for general-domain video-language pre-training.
ACO~\cite{Zhang2022ACO} and SelfD~\cite{Zhang2022SelfD} are the only two that collect 120 and 100 hours of driving videos from YouTube, respectively, to pre-train an encoder for policy learning (Details in \cref{sec:related-web-driving-videos}). In contrast, we exhaustively mine driving videos from YouTube and construct the largest driving video datasets publicly available, accumulating over 1700 hours. Besides, our videos are paired with descriptions and command labels which can be used for broader applications such as language-guided autonomous driving~\cite{Chen2023E2ESurvey, li2023_datasetsurvey, Sima2023DriveLM, Zhou2024ELM}.

\section{\datasuitename Dataset}
\label{supp:dataset}

Our data suite, \datasuitename, the \textit{largest} public driving dataset to date, contains 2059 hours of driving video along with diverse text conditions, including \textit{contexts} and \textit{commands}. In this section, we detail the YouTube video collection process (\cref{supp:collect}), language annotation method (\cref{supp:anno} for \youtubesplitename and \cref{supp:public_dataset} for other public datasets), more examples and analysis to illustrate the diversity of \datasuitename (\cref{supp:our_dataset_anl} and \cref{supp:our_div}).

\begin{figure*}[!t]
    \centering
    \includegraphics[height=1.15\textwidth]{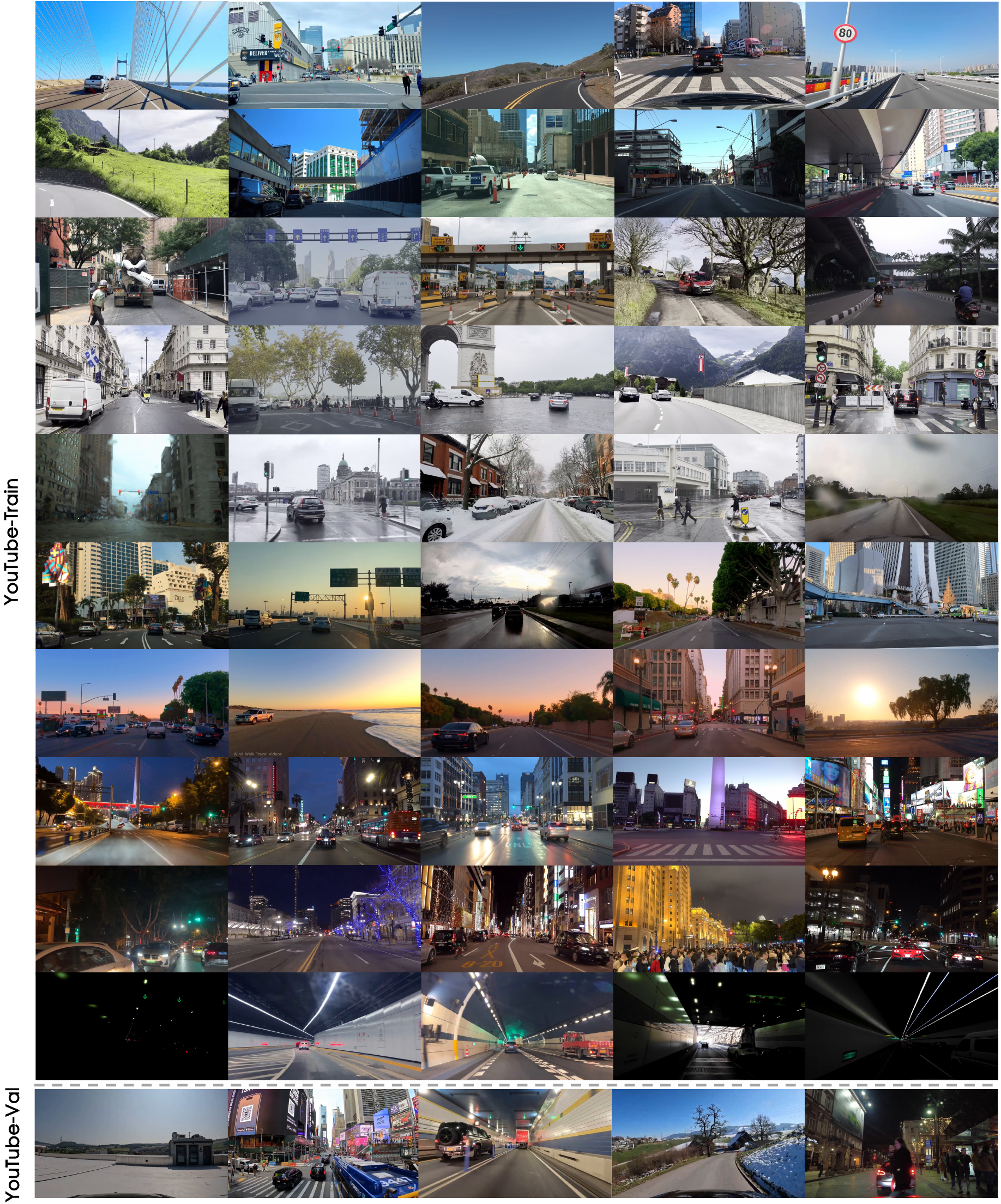}    
    \caption{\textbf{Diverse video samples in \youtubesplitename}. We only showcase certain frames from videos due to space limits. \youtubesplitename covers a wide spectrum of diversity in multiple axes, including geographic locations, traffic scenarios, time periods, weather conditions, \etc. 
    We strictly construct the Train/Val split from 
    different YouTubers for zero-shot evaluation.
    }
    \label{fig:youtube_videos}
\end{figure*}

\subsection{\youtubesplitename}
\label{supp:our_dataset}

\subsubsection{Data Collection}
\label{supp:collect}

\paragraph{Data Acquisition.} We first search for videos of driving tours on YouTube and select 43 video uploaders worldwide, \ie, YouTubers, who continuously post high-quality driving videos.
We further check the quality of videos from these YouTubers in terms of resolution, frame rate, scene transition frequency, \etc, resulting in 2139 high-quality front-view driving videos. 
We take all videos from 3 selected YouTubers as the validation set, including
\cmtt{Pete Drives USA}, \cmtt{KenoVelicanstveni}, and \cmtt{Driving Experience}, while the other videos are used for training. 
We illustrate the diversity of the \youtubesplitename in \cref{fig:youtube_videos}.

\paragraph{Format Conversion.} To simplify the data usage for training both image and video models,
We pre-process all videos into sets of consecutive frames in image format using \cmtt{decord} and \cmtt{opencv} packages. We sample videos with 
resolutions no less than 720p (\eg, 1280$\times$720 for $16:9$ videos) at 10Hz.

\paragraph{Data Cleaning.} 
To ensure the quality of our dataset, we exclude non-driving frames which are commonly shown in each video and introduce unwanted noise. Specifically, we discard the first 90 seconds and the last 30 seconds for most videos to remove the channel introduction at the beginning and the subscription reminder at the end.
For YouTubers with longer video introductions, we discard the first 180 or 300 seconds from their videos. We further detect and remove black frames and transition frames with the help of vision-language models. 
We first search for frames with phrases like \cmtt{words, watermark, dark night, dark street,} and \cmtt{blur} in their BLIP-2~\cite{Li2023BLIP2} -generated contexts, followed by the manual quality check 
to determine their removal. For details on BLIP-2 descriptions, please refer to \cref{supp:anno}.

\begin{figure}[tb!]
    \centering
    \includegraphics[width=.8\linewidth]{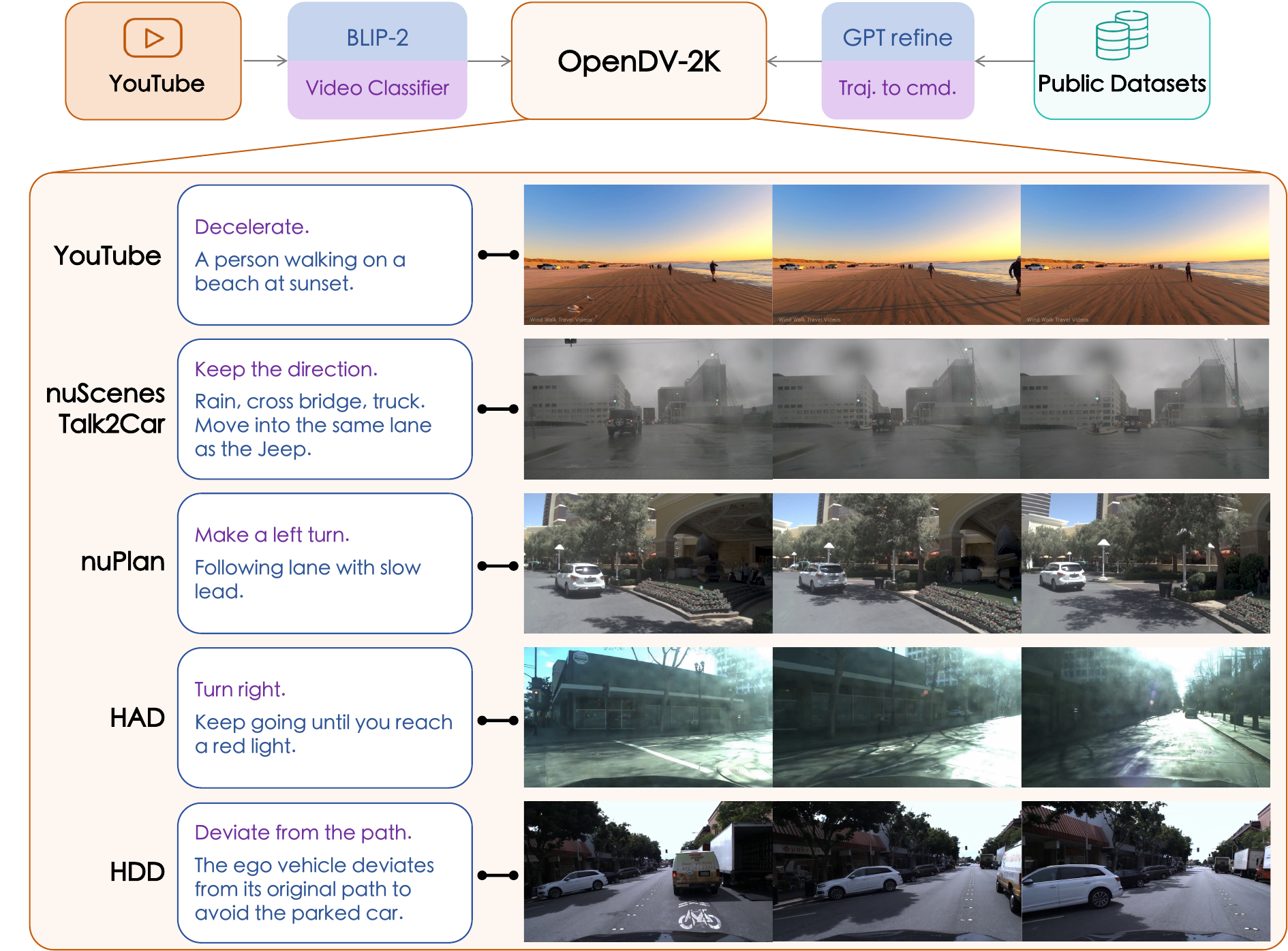}
    \caption{\textbf{Examples of language annotations for different data sources in \datasuitename}.
    We unify the paired text as  
    \textcolor{violet}{command} and \textcolor{DarkBlue}{context} for all data sources after careful pre-processing. The \textcolor{violet}{command} represents the action of the ego vehicle, whereas the \textcolor{DarkBlue}{context} covers various aspects of information in the driving scenario.
    For details on how to merge public driving datasets, please refer to \cref{supp:public_dataset}.}
    \label{fig:our_anno}
\end{figure}

\subsubsection{Language Annotation}
\label{supp:anno}

Our \youtubesplitename possesses two types of annotations, frame descriptions (contexts) and ego-driver commands. The context aims to benefit text-to-image learning, helping the model understand the concepts of open-world objects and scenarios, whereas the command is designed to correlate the future predictions with ego actions and further enables the language as control signals.
We show some examples in \cref{fig:our_anno} and introduce the annotation method below.

\paragraph{Frame Descriptions (Contexts).} We leverage the established BLIP-2~\cite{Li2023BLIP2} to describe the main objects or scenarios in each frame with the following prompt. The language annotations are also used in data cleaning, as mentioned in \cref{supp:collect}. 

{
\captionsetup{type=table}
\begin{tcolorbox}[enhanced, breakable,
                  colback=gray!10,%
                  colframe=black,%
                  width=\linewidth,
                  arc=1mm, auto outer arc,
                  boxrule=0.5pt,
                 ]
\small
\cmtt{\textcolor{blue}{Prompt} = "}Question: Describe the image of a driving scenario concisely. Answer: \cmtt{"}
\end{tcolorbox}
\vspace{-10pt} 
\captionof{table}{\textbf{BLIP-2 Prompt for generating context of each frame}.}
\label{box:blip}
}

\paragraph{Driver Commands.} Similar to the conventional behavior planning approach~\cite{Ramanishka2018HondaHDD}, we classify the commands for ego vehicle into 13 categories,
\ie, \{\cmtt{forward, intersection passing, left turn, right turn, left lane change, right lane change, left lane branch, right lane branch, crosswalk passing, rail passing, merge, U-Turn, stop/decelerate, deviate}\}. 
We train an action model based on optical flow to annotate the command for the unlabeled YouTube dataset.
Specifically, 
we leverage the pre-trained GMFlow~\cite{Xu2022GMFlow, Xu2023Unifying} to extract optical flow between adjacent frames of a driving video sequence. 
Taking as input both the optical flow and its distance map~\cite{Zheng2021DistanceMap},
we train a ResNet-18~\cite{He2016ResNet18} to classify the action of each 4s video clip.
The training is conducted on the merged dataset of Honda-HDD-Action and Honda-HDD-Cause~\cite{Ramanishka2018HondaHDD}, which provides specified action annotations.
For each type of action, we match it with multiple expressions to enrich language understanding. During training, we randomly select one text from the matched caption set for each action.
The dictionary for paraphrasing is shown in \cref{box:cmd} below.

\vspace{5pt}
{
\vspace{-5pt}
\captionsetup{type=table}
\begin{tcolorbox}[enhanced, breakable,
                  colback=gray!10,%
                  colframe=black,%
                  width=\linewidth,
                  arc=1mm, auto outer arc,
                  boxrule=0.5pt,
                 ]
\small
\cmtt{\textcolor{blue}{command\_caption\_dict} = \{ \\
    \footnotesize
    \hspace{-1cm}
    \begin{longtable}{lllll}
        0: [ 
        & "\textrm{Move forward.}", 
        & "\textrm{Move steady.}", 
        & "\textrm{Go forward.}", \\
        & "\textrm{Go straight.}", 
        & "\textrm{Proceed.}", 
        & "\textrm{Drive forward.}", \\
        & "\textrm{Drive straight.}", 
        & "\textrm{Drive steady.}", 
        & "\textrm{Keep the direction.}", \\
        & "\textrm{Maintain the direction.}" 
        & & & ], \\
        \\
        1: [ 
        & "\textrm{Pass the intersection.}", 
        & "\textrm{Cross the intersection.}", 
        & "\textrm{Traverse the intersection.}", \\
        & "\textrm{Drive through the intersection.}", 
        & "\textrm{Move past the intersection.}", 
        & "\textrm{Pass the junction.}", \\
        & "\textrm{Cross the junction.}", 
        & "\textrm{Traverse the junction.}", 
        & "\textrm{Drive through the junction.}", \\
        & "\textrm{Move past the junction.}", 
        & "\textrm{Pass the crossroad.}", 
        & "\textrm{Cross the crossroad.}", \\
        & "\textrm{Traverse the crossroad.}", 
        & "\textrm{Drive through the crossroad.}", 
        & "\textrm{Move past the crossroad.}"
        &], \\
        \\
    2: [ 
        & "\textrm{Turn left.}", 
        & "\textrm{Turn to the left.}", 
        &"\textrm{Make a left turn.}", \\
        & "\textrm{Take a left turn.}", 
        & "\textrm{Turn to the left.}", 
        & "\textrm{Left turn.}", \\
        & "\textrm{Steer left.}", 
        & "\textrm{Steer to the left.}"
        & & ], \\
        \\
    3: [ 
        & "\textrm{Turn right.}",
        & "\textrm{Turn to the right.}", 
        & "\textrm{Make a right turn.}", \\
        & "\textrm{Take a right turn.}", 
        & "\textrm{Turn to the right.}",
        & "\textrm{Right turn.}", \\
        & "\textrm{Steer right.}", 
        & "\textrm{Steer to the right.}"
        & & ], \\
        \\
    4: [ 
        & "\textrm{Make a left lane change.}", 
        & "\textrm{Change to the left lane.}", 
        & "\textrm{Switch to the left lane.}", \\
        & "\textrm{Shift to the left lane.}", 
        & "\textrm{Move to the left lane.}"
        & & ], \\
        \\
    5: [ 
        & "\textrm{Make a right lane change.}",
        & "\textrm{Change to the right lane.}", 
        & "\textrm{Switch to the right lane.}", \\
        & "\textrm{Shift to the right lane.}", 
        & "\textrm{Move to the right lane.}"
        & & ], \\
        \\
    6: [ 
        & "\textrm{Go to the left lane branch.}", 
        & "\textrm{Take the left lane branch.}", 
        & "\textrm{Move into the left lane branch.}", \\
        & "\textrm{Follow the left lane branch.}", 
        & "\textrm{Follow the left side road.}"
        & & ], \\
        \\
    7: [ 
        & "\textrm{Go to the right lane branch.}", 
        & \textrm{Take the right lane branch.}", 
        & "\textrm{Move into the right lane branch.}", \\
        & "\textrm{Follow the right lane branch.}", 
        & "\textrm{Follow the right side road.}"
        & &], \\
        \\
    8: [
        & "\textrm{Pass the crosswalk.}", 
        & "\textrm{Cross the crosswalk.}", 
        & "\textrm{Traverse the crosswalk.}", \\
        & "\textrm{Drive through the crosswalk.}", 
        & "\textrm{Move past the crosswalk.}", 
        & "\textrm{Pass the crossing area.}", \\
        & "\textrm{Cross the crossing area.}", 
        & "\textrm{Traverse the crossing area.}", 
        & "\textrm{Drive through the crossing area.}", \\
        & "\textrm{Move past the crossing area.}"
        & & & ], \\
    \\
    9: [
        & "\textrm{Pass the railroad.}", 
        & "\textrm{Cross the railroad.}", 
        & "\textrm{Traverse the railroad.}", \\
        & "\textrm{Drive through the railroad.}", 
        & "\textrm{Move past the railroad.}", 
        & "\textrm{Pass the railway.}", \\
        & "\textrm{Cross the railway.}", 
        & "\textrm{Traverse the railway.}", 
        & "\textrm{Drive through the railway.}", \\
        & "\textrm{Move past the railway.}"
        & & & ], \\
    \\
    10: [
        & "\textrm{Merge.}", 
        & "\textrm{Merge traffic.}", 
        & "\textrm{Merge into traffic.}", \\
        & "\textrm{Merge into the traffic.}", 
        & "\textrm{Join the traffic.}", 
        & "\textrm{Merge into the traffic flow.}", \\
        & "\textrm{Join the traffic flow.}", 
        & "\textrm{Merge into the traffic stream.}", 
        & "\textrm{Join the traffic stream.}", \\
        & "\textrm{Merge into the lane.}",
        & & & ], \\
    \\
    11: [
        & "\textrm{Make a U-turn.}",
        & "\textrm{Make a 180-degree turn.}",
        & "\textrm{Turn 180 degree.}", \\
        & "\textrm{Turn around.}",
        & "\textrm{Drive in a U-turn.}"
        & & ], \\
    \\
    12: [
        & "\textrm{Stop.}", 
        & "\textrm{Halt.}", 
        & "\textrm{Decelerate.}", \\
        & "\textrm{Slow down.}", 
        & "\textrm{Brake.}"
        & & ], \\
    \\
    13: [
        & "\textrm{Deviate.}", 
        & "\textrm{Deviate from the path.}", 
        & "\textrm{Deviate from the lane.}", \\
        & "\textrm{Change the direction.}", 
        & "\textrm{Shift the direction.}"
        & & ] \\
\end{longtable}
\} 
}
\end{tcolorbox}
\vspace{-10pt} 
\captionof{table}{\textbf{Paraphrasing dictionary for command generation}. Each index corresponds to one of the 13 actions inferred by the classifier.}
\label{box:cmd}
}

\vspace{10pt}
\subsubsection{Analyses Methods}
\label{supp:our_dataset_anl}

In this section, we elaborate on the means of data analysis for \youtubesplitename. The analysis results are reported in the main paper and \cref{supp:our_div}.

\paragraph{Geographic Diversity Analysis.} We take GPT-3.5-turbo~\cite{Ouyang2022InstructGPT} to infer the geographic information of each video from its title. We also apply handmade rules to post-process the results from GPT-3.5-turbo to deal with multiple aliases of one city or one country. 
The prompts 
are shown in \cref{box:geo} where \cmtt{\small\textcolor{blue}{title}} denotes the video title to be inferred.
For simplicity, we assume that all clips of a video are taken in the same place.
For videos with multiple inferred locations, we assume that all clips included in that video are uniformly distributed in these locations. For a video composed of $M$ clips with $N$ inferred locations, we assume there are $\frac{M}{N}$ clips taken in each site.

{
\captionsetup{type=table}
\begin{tcolorbox}[enhanced, breakable, 
                  colback=gray!10,%
                  colframe=black,%
                  width=\linewidth,
                  arc=1mm, auto outer arc,
                  boxrule=0.5pt,
                 ]
\small
{\cmtt{\textcolor{blue}{\textbf{Messages}}
 = [ 

\bigskip

 \{
"role": "system",
"content": f"""}
You are a helpful assistant, who is a geography expert and is also good at recognizing different languages.
\cmtt{""" \},}

\bigskip
\cmtt{\{
"role": "user",
"content": f"""} 
Try to infer in which city or state a video is taken from its title. Please answer the city name, the state name, and the country name in English respectively and briefly, in the following form: $\backslash$

\bigskip
``Country: \{\{the name of the country\}\}

\smallskip
State: \{\{the name of the state or the province\}\}

\smallskip
City: \{\{the name of the city\}\}". $\backslash$

\bigskip
If something cannot be inferred, fill the corresponding blank with ``N/A". If there is more than one city in the video, first check if all the answers are valid, i.e. the name of cities, instead of the names of districts or towns. If there are multiple cities after checking the validity, use ``," to separate different cities. $\backslash$

\bigskip
You should also try to infer the state or province where the cities belong and fill the answer into the blank of ``State". Note that you must infer the country where the video might be taken. Moreover, please discard meaningless words like ``city", ``country", ``province" or ``state" when filling in the blanks. $\backslash$

\bigskip
The title of the video is as follows: 
\{\cmtt{\textcolor{blue}{title}\}"""}\}]
}

\end{tcolorbox}
\vspace{-10pt} \captionof{table}{\textbf{Prompt for geographic inference of videos}.}
\label{box:geo}
}

\paragraph{Scenario Diversity Analysis.} For scene analysis, we visualize the frequency of different scenes in frame descriptions generated in \cref{supp:anno}. For analyses on weather and time period, we observe that some language hints such as ``foggy'' and ``night'' are often present in videos' titles, thus we prompt GPT-3.5-turbo~\cite{Ouyang2022InstructGPT} to infer the weather and photographed period of the video from its title. 
The prompt is shown in \cref{box:scene}
where \cmtt{\small\textcolor{blue}{title}} denotes the video title to be inferred.

{
\captionsetup{type=table}
\begin{tcolorbox}[enhanced, breakable,
                  colback=gray!10,%
                  colframe=black,%
                  width=\linewidth,
                  arc=1mm, auto outer arc,
                  boxrule=0.5pt,
                 ]

\small
{\cmtt{\textcolor{blue}{\textbf{Messages}}}
 = [

\bigskip
\cmtt{
 \{
"role": "system",
"content": f"} You are a helpful assistant, who has a good command of multiple languages.
\cmtt{"\},}

\bigskip
\cmtt{\{
"role": "user",
"content": f"""} Try to infer in which weather and period a video is taken from its title. Please answer the weather and period in English respectively and briefly, in the following form: $\backslash$

\medskip
``Weather: \{\{the weather\}\}

\smallskip
Period: \{\{the period\}\}". $\backslash$

\bigskip
If something cannot be inferred, fill the corresponding blank with ``N/A". The weather must be one of the following: ``sunny", ``rainy", ``foggy", ``snowy",   ``cloudy", ``storm". The period must be one of the following: ``daytime", ``dusk", ``dawn", ``nighttime". $\backslash$

\bigskip
The title of the video is as follows: 
\{\cmtt{\textcolor{blue}{title}\}"""}\}]
}

\end{tcolorbox}
\vspace{-10pt}
\captionof{table}{\textbf{Prompt for weather and time period inference of videos}.}
\label{box:scene}
}

\subsubsection{Diversity Highlights}
\label{supp:our_div}

\paragraph{Geographic Distribution.} 
As indicated by the human-refined GPT inference results,
YouTube videos are taken from over 244 cities in more than 40 countries, covering considerably more areas than any existing public driving datasets, as shown in Tab.~\red{1} and Fig.~\red{2} in the main paper.
Note that the result is still \textit{underestimated} since the geographic information may not be included in the title for some videos and cannot be inferred.
Taking the two most popular areas as an example, \youtubesplitename contains 36.4M clips in the US, covering 40 out of 50 states, and 12.9M clips in China, covering 26 out of 34 provinces. 
Moreover, to test the zero-shot performance of a model in its unseen locations, our YouTube-Val subset contains videos from 3 countries that are not included in YouTube-Train, \ie, \cmtt{Bosna i Hercegovina, Denmark,} and \cmtt{Hungary}. There are also videos from 1 state of the US unseen in YouTube-Train, \ie, \cmtt{Maine}.

\paragraph{Camera Settings.} 
Considering that the online videos are sourced from different YouTubers around the globe, our dataset enjoys high diversity in photography equipment, leading to plentiful color settings, camera intrinsic parameters, and camera poses.
For instance, a front-view video on a double-deck bus (see the second left picture in the last row of \cref{fig:youtube_videos}) is provided in our YouTube-Val subset while no similar cases are included in the YouTube-Train subset.

\paragraph{Scenarios.} We claim that there is sufficient data of diverse driver actions, weather conditions, photographed periods, and scenes in our \youtubesplitename.
Results are shown in \cref{tab:action_distribution}, \cref{tab:weather_distribution}, \cref{tab:period_distribution}, and \cref{fig:scene_distribution}, respectively.
Note that according to the analysis process in \cref{supp:our_dataset_anl}, the diversity of scenarios in our dataset is \textit{estimated} values since not all videos provide weather and filming periods in their titles.

\begin{table*}[!t]
\begin{minipage}[b]{0.43\textwidth}
    \captionsetup{type=figure}
    \centering
    \includegraphics[width=.96\linewidth]{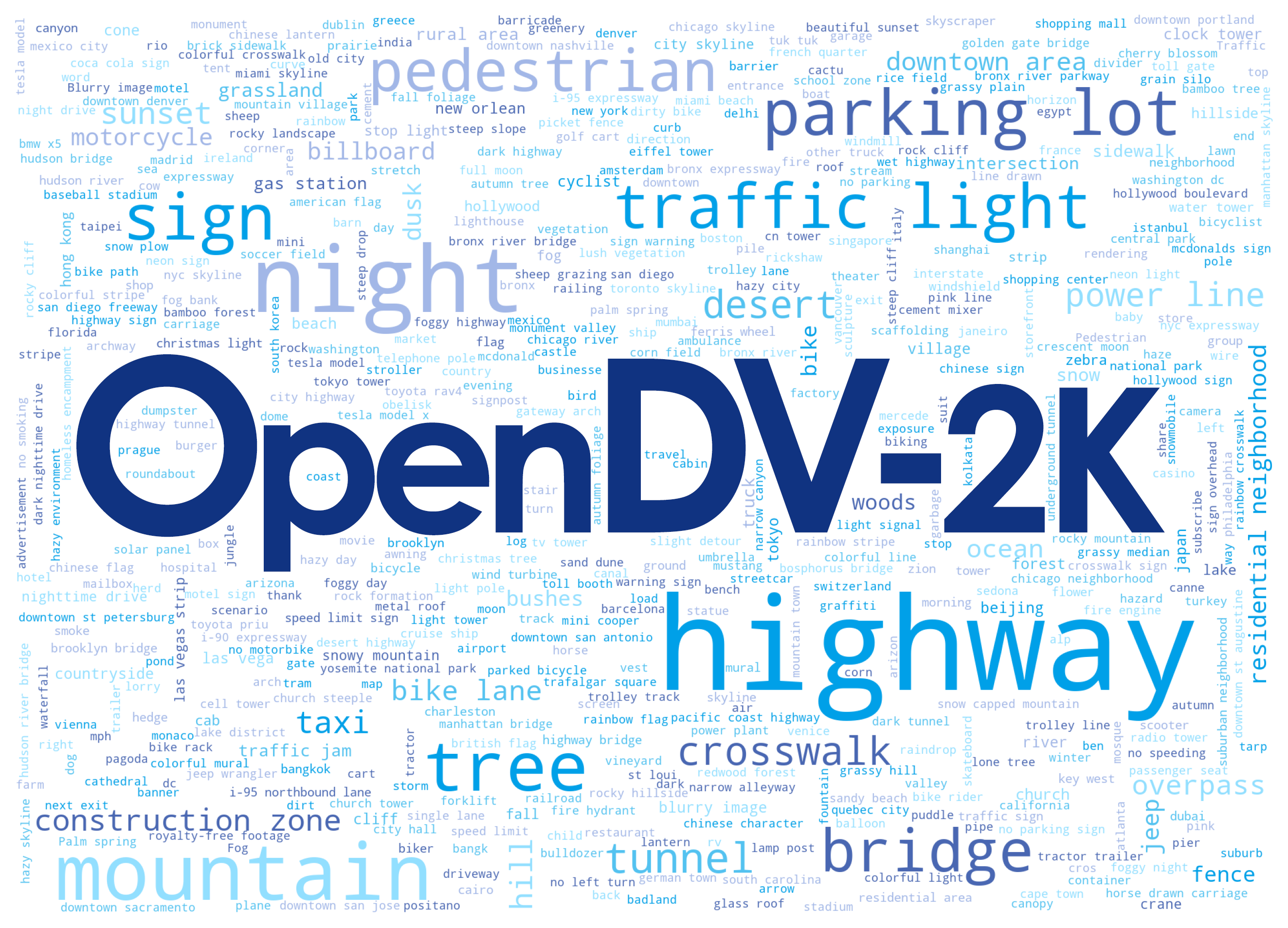}
    \vspace{-5pt}
    \captionof{figure}{\textbf{Word cloud of frame descriptions for \datasuitename}. Only the top 500 most frequently mentioned objects, agents, or scenarios are included in the word cloud.}
    \label{fig:scene_distribution}
\end{minipage}
\hfill
\begin{minipage}[b]{0.55\textwidth}
    \centering
    \scriptsize
    \begin{tabular}{c|ccccccc}
        \toprule
        \makecell{Driver\\Action} & Forward & Stop & \makecell{Left\\Turn} & \makecell{Right\\Turn} & U-Turn & \makecell{Lane\\Change} & \makecell{Intersection\\Passing} \\
        \midrule
        \makecell{Estimated\\Proportion} & 81.39$\%$ & 8.85$\%$ & 1.89$\%$ & 1.81$\%$ & 0.27$\%$ & 0.30$\%$ & 5.49$\%$ \\
        \bottomrule
    \end{tabular}
    \vspace{-5pt}
    \caption{\textbf{Driver action distribution of \youtubesplitename}.}
    \label{tab:action_distribution}
    \vspace{.3cm}
    \centering
    \begin{tabular}{c|cccc}
        \toprule
        Period & Daytime & Dawn & Dusk & Nighttime  \\
        \midrule
        \makecell{Estimated\\ Frame Count} & 54M & 425K & 2M & 4M \\
        \bottomrule
    \end{tabular}
    \vspace{-5pt}
    \caption{\textbf{Time period distribution of \youtubesplitename}.}
    \label{tab:period_distribution}
    \vspace{.3cm}
    \centering
    \begin{tabular}{c|cccccc}
        \toprule
        Weather & Normal & Rainy & Cloudy & Foggy & Snowy & Storm \\
        \midrule
        \makecell{Estimated\\ Frame Count} & 58M & 690K & 503K & 284K & 503K & 117K \\
        \bottomrule
    \end{tabular}
    \vspace{-5pt}
    \caption{\textbf{Weather distribution of \youtubesplitename}.}
    \label{tab:weather_distribution}
\end{minipage}
\end{table*}

\paragraph{Corner Cases.} YouTube videos also contain corner cases and safety-critical cases. Several special cases from \youtubesplitename are given in \cref{fig:youtube_videos}, \eg, dark tunnels with limited lighting (the leftmost and the rightmost in the 2nd row from bottom), intersections crowded with numerous pedestrians during nighttime (the 2nd right in the 3rd row from bottom), beaches at sunset (the 2nd left in the 5th row from bottom), rooftop (the leftmost in the last row), and videos captured with raindrops on the camera lens (the rightmost in the 5th row from top).

\subsection{Merged Public Datasets}
\label{supp:public_dataset}

Though the annotations in the \youtubesplitename are on a large scale, annotations are subject to limited patterns. \textit{Contexts} from BLIP-2 follow certain syntax while \textit{commands} are generated by the paraphrase dictionary. To provide more diverse expressions of contexts and commands, we merge annotations and sensor data from existing public datasets after converting their labels into complete sentences with correct grammar and format.

\subsubsection{Contexts Generation}
\label{supp:desc}

\noindent\textbf{nuScenes \& nuPlan.} Contexts are directly inherited from the scenario description of its belonging scenario in nuScenes~\cite{Caesar2019nuScenes} or nuPlan~\cite{Caesar2021nuPlan}.

\smallskip
\noindent\textbf{ONCE.} In the metadata of ONCE~\cite{Mao2021ONCE}, weather condition and filming time period are provided. These annotations are directly inherited in \splitname{ONCE} as contexts.

\smallskip
\noindent\textbf{Honda-HAD.} Diverse contexts are generated by refining and paraphrasing driving events provided by Honda-HAD~\cite{Kim2019HondaHAD}. The prompt for refinement and paraphrasing is as follows.

{
\captionsetup{type=table}
\begin{tcolorbox}[enhanced, breakable,
                  colback=gray!10,%
                  colframe=black,%
                  width=\linewidth,
                  arc=1mm, auto outer arc,
                  boxrule=0.5pt,
                 ]
\small
\cmtt{\textcolor{blue}{\textbf{Messages}}
 = [ 

\bigskip
 \{
"role": "system",
"content": f"""} You are a helpful assistant.
\cmtt{""" \},}

\bigskip
\cmtt{\{
"role": "user", 
"content": f"""}Generate \cmtt{\{\textcolor{blue}{NUM\_GEN}\}} descriptions with exactly the same meanings as the following reference sentence, REF: \{\cmtt{\textcolor{blue}{current\_caption}}\}. $\backslash$

\bigskip
          Please write these sentences concisely in diverse ways, and try to use common and simple words if possible. $\backslash$

\bigskip
          Each generated sentence denotes a short description of noteworthy elements (e.g. pedestrians, traffic lights, cars) in this driving scenario. $\backslash$

\bigskip
          There might be some typos, grammar errors, or unnatural expressions in the REF sentence, and you might need to correct these issues in the generated sentences. Each generated sentence should be correct in grammar and spelling, easy to understand, in natural and smooth expression.
          All sentences have the same meaning with the reference sentence REF, and the only difference is the wording. $\backslash$

\bigskip

          Your complete response is only a python list including \{\textcolor{blue}{\cmtt{NUM\_GEN}}\} strings (No other text needed), 
          each one is an example sentence with an identifier \texttt{`}\cmtt{$\backslash$n}\texttt{'} in the end.
\cmtt{"""\}]}

\end{tcolorbox}
\vspace{-10pt} \captionof{table}{\textbf{Prompt for paraphrasing contexts in Honda-HAD dataset}.}
\label{box:had-desc}
}

\subsubsection{Commands Annotation}

\noindent\textbf{nuScenes \& nuPlan.} Since vehicle trajectory is given in nuScenes~\cite{Caesar2019nuScenes} and nuPlan~\cite{Caesar2021nuPlan}, ego-vehicle commands can be easily calculated from trajectories by mathematical methods. After commands are generated, we can refer to \cref{box:cmd} to provide diverse expressions of driver commands.

\smallskip
\noindent\textbf{Talk2Car.} Talk2Car~\cite{Deruyttere2019Talk2Car} provides texts of possible human intentions for each scene in nuScenes. These annotations are inherited after they are refined by GPT-3.5-turbo to be grammatically correct and in appropriate formats. The prompt used for refinement is as follows.

{
\captionsetup{type=table}
\begin{tcolorbox}[enhanced, breakable,
                  colback=gray!10,%
                  colframe=black,%
                  width=\linewidth,
                  arc=1mm, auto outer arc,
                  boxrule=0.5pt,
                 ]
\small
\cmtt{\textcolor{blue}{\textbf{Messages}}
 = [ 

\bigskip
 \{
"role": "system",
"content": f"""} You are a helpful assistant.
\cmtt{""" \},}

\bigskip
\cmtt{\{
"role": "user", 
"content": f"""} Please correct the capitalization and punctuation issues in this sentence: ``\{\cmtt{\textcolor{blue}{current\_caption}}\}". The original characters and words should be exactly the same without any changes. Do not add quotation marks.
\cmtt{"""\}]}

\end{tcolorbox}
\vspace{-10pt} \captionof{table}{\textbf{Prompt for refining texts in Talk2Car dataset}.}
\label{box:talk2car}
\vspace{10pt}
}

\smallskip
\noindent\textbf{ONCE.} Behaviours of the ego-vehicle can be obtained from the change in camera pose provided in ONCE~\cite{Mao2021ONCE}. They are further converted to natural language using \cref{box:cmd}.

\smallskip
\noindent\textbf{Honda-HAD.} Since driver behaviours are not directly provided in Honda-HAD~\cite{Kim2019HondaHAD}, we implement the video classifier trained in \cref{supp:anno} and refer to \cref{box:cmd} to generate ego-vehicle behaviours. Moreover, Honda-HAD does provide sufficient driving advice for each scene. We use GPT-3.5-turbo to refine and paraphrase these annotations so that diverse expressions are contained in our \datasuitename. Prompts for GPT-3.5-turbo are as follows.

{
\captionsetup{type=table}
\begin{tcolorbox}[enhanced, breakable,
                  colback=gray!10,%
                  colframe=black,%
                  width=\linewidth,
                  arc=1mm, auto outer arc,
                  boxrule=0.5pt,
                 ]

\small
\cmtt{\textcolor{blue}{\textbf{Messages}}
 = [ 

\medskip
 \{
"role": "system",
"content": f"""} You are a helpful assistant.
\cmtt{""" \},}

\bigskip
\cmtt{\{
"role": "user", 
"content": f"""}Generate \cmtt{\{\textcolor{blue}{NUM\_GEN}\}} driving commands with exactly the same meanings as the following sentence: \cmtt{\{\textcolor{blue}{current\_caption}\}}. $\backslash$

\bigskip
          Please write these sentences concisely in diverse ways, and try to use common and simple words if possible.
          Remember all sentences have the same meaning, which is an instruction or intention for the planning of the ego vehicle. $\backslash$

\bigskip
          Your complete response is only a python list including \cmtt{\{\textcolor{blue}{NUM\_GEN}\}} strings (No other text needed), 
          each one is an example sentence with an identifier \texttt{`}\cmtt{$\backslash$n}\texttt{'} in the end.
\cmtt{"""\}]}

\end{tcolorbox}
\vspace{-10pt} \captionof{table}{\textbf{Prompt for paraphrasing command annotations from driving advice in Honda-HAD dataset}.}\label{box:had-cmd}
\vspace{15pt}
}

\smallskip
\noindent\textbf{Honda-HDD-Action.} Honda-HDD-Action~\cite{Ramanishka2018HondaHDD} contains 104 hours of videos with corresponding labels of driving commands. Since some clips begin with transitions from a completely green frame, we remove the first 30 frames from all clips.
Moreover, driving events in videos with a duration too long might be inconsistent with human-annotated behaviors. Therefore, we have to discard all video clips longer than 20 seconds. Meanwhile, since in the training stage, our model takes videos no shorter than 4 seconds as input, we also remove all videos shorter than 4 seconds. Only 32 hours of videos are left after this cleaning process. For the remaining clips, we directly use the labels as driver commands and use \cref{box:cmd} to generate command texts.

\smallskip
\noindent\textbf{Honda-HDD-Cause.} There are 12 hours of videos in Honda-HDD-Cause~\cite{Ramanishka2018HondaHDD}, as well as corresponding human-annotated driving behaviors and human explanations. Similar to Honda-HDD-Action, we apply the same cleaning process on Honda-HDD-Cause, with about 1 hour of cleaned driving videos preserved. To align with \youtubesplitename, we convert these videos into frame sets by sampling the sensor videos at 10Hz. For command annotations, causal explanations in the form of phrases in the original dataset are inherited after refining and paraphrasing by GPT-3.5-turbo. The prompts used are as follows.

{
\captionsetup{type=table}
\begin{tcolorbox}[enhanced, breakable,
                  colback=gray!10,%
                  colframe=black,%
                  width=\linewidth,
                  arc=1mm, auto outer arc,
                  boxrule=0.5pt,
                 ]
\small

\cmtt{\textcolor{blue}{\textbf{elements}}= "}sign, congestion, traffic light, pedestrian, parked car\cmtt{"}

\bigskip
\cmtt{\textcolor{blue}{\textbf{Messages}}
 = [ 
 
\bigskip
 \{
"role": "system",
"content": f"""} \textrm{You are a helpful AI driving assistant, 
        who gives commands to the ego vehicle in natural language for safe driving.}  $\backslash$

        \bigskip
        You are provided with one of the following elements of the driving scenario, namely, \cmtt{\{\textcolor{blue}{elements}\}}. Based on the given element, produce a driving command indicating either \texttt{`}stop\texttt{'} or \texttt{`}deviate\texttt{'} to the ego vehicle.  Specifically, sign, congestion, traffic light, crossing vehicle, and pedestrian simulates a \texttt{`}stop\texttt{'} command, 
        and only the parked car leads to a \texttt{`}deviate\texttt{'} command. $\backslash$

        \bigskip
        You should write \cmtt{\{\textcolor{blue}{NUM\_GEN}\}} fluent, concise, and diverse sentences for each command, using common and simple words.
        Half of these sentences are descriptions of the action of the ego vehicle (or driver), and the other half should be imperative sentences. 
        All sentences should have the same meaning, and the only difference is the wording.
\cmtt{""" \},}

\bigskip
\cmtt{\{
"role": "user", 
"content": \textcolor{blue}{current\_caption}\}]}

\end{tcolorbox}
\vspace{-10pt}
\captionof{table}{\textbf{Prompt for refining driving commands in Honda-HDD-Cause dataset}.}
\label{box:had}
}

\bigskip
\section{Implementation Details of \modelname}
\label{supp:implementation}

\subsection{Model Design}
\label{supp:model}

\subsubsection{GenAD}
\modelname is built upon 2.7B SDXL~\cite{Podell2023SDXL}, which is a large-scale text-to-image generation model. We first fine-tune it in the first stage to transfer its domain knowledge to driving view synthesis. After that, we freeze the original blocks in the denoising UNet, and interleave them with our proposed temporal reasoning blocks, in total 2.5B, to allow for modeling on video sequences in video prediction pre-training. 
Following the original SDXL, the language conditions are encoded by two frozen CLIP variants with 817M parameters, namely, CLIP ViT-L~\cite{Radford2021CLIP} and OpenCLIP ViT-bigG~\cite{Cherti2023OpenClip}, and the projection between the pixel space and latent space is performed by a pre-trained autoencoder with 83.7M parameters. As a result, \modelname has 5.9B parameters in total. The computational complexity is 5.27 TFLOPs.

\subsubsection{Extension on Action-condition Prediction}
Besides the text conditions and past-frame conditions, we introduce the future trajectory of the ego vehicle as an additional condition signal to guide the denoising and therefore control the future imagination. We implement it by transforming the low-dimensional future waypoints into high-dimensional continuous embeddings~\cite{tancik2020Fourier}, then projecting it with a zero-initialized linear layer into the same dimension with the text conditions $\rvc$. 
With zero initialization, the knowledge of future trajectory could be gradually injected into the model through the conditional cross-attention layer alongside $\rvc$, avoiding disturbing the learned prior on other conditions in the first place.  
It further controls the future simulation to be consistent with the ego intentions. Here the conditional future trajectory includes 6 waypoints at 2Hz.

\subsubsection{Extension on Planning}
Since \modelname is capable of predicting reasonable futures given past observations, it encodes past frames in a meaningful way to guide the denoising of future frames. Therefore, we take the pre-trained \modelname as a strong feature extractor to obtain the spatiotemporal representations from past frames for downstream policy learning.
We only utilize the encoder part of \modelname's denoising UNet to extract intermediate semantic features rather than acquiring the noise from the decoder part.
Specifically, given the past two frames and a high-level command generated in the same way as in~\cite{Hu2023UniAD, hu2022stp3}, the frozen \modelname encoder extracts the spatiotemporal features, which are passed to a randomly initialized multi-layer perceptron (MLP) to project them into the future trajectory of ego vehicle.
The MLP is composed of 6 linear layers and 5 ReLU activations, containing only 0.8M parameters in total.
The first two linear layers downsample the features channel-wise, then the features of two frames are concatenated in channel dimension and further downsampled by the third linear layer. After that, the features are average-pooled in spatial dimensions, and the resulting vector is projected to the future trajectory, which is composed of 6 waypoints at 2Hz (3s), by the last two linear layers of the MLP.

\subsection{Training Details}
\label{supp:training}

\modelname is trained in two phases, \ie, image domain transfer and video prediction pre-training.
In the first stage,
we fine-tune the pre-trained SDXL on per-image denoising 
with 2.7B trainable parameters of its denoising UNet.
It is trained on 
65.1M image-text pairs of \datasuitename. 
Each text condition is unified as ``command, context".
For some commands and contexts that are originally labeled for video sequences rather than static images, we simply associate them with all image frames included in that video sequence.
We train the model for 300K iterations on 32 
GPUs with a total batch size of 256 with AdamW~\cite{Loshchilov2018AdamW}.
We linearly warm up the learning rate for $10^4$ steps in the beginning then keep it constant at $1.25 \times 10^{-6}$. The default GPUs in most of our experiments are NVIDIA Tesla A100 devices unless otherwise specified.

In the second stage, we train the model on video-level denoising using video-text pairs lifting it to predict the future iteratively during inference.
For compute efficiency, we freeze all blocks of the fine-tuned image model and only optimize our introduced temporal reasoning blocks, resulting in 2.5B trainable parameters in this stage.
To maximize the data efficiency for constructing video clips, we take each frame of a 10Hz YouTube video as a starting frame to form a 4s training sequence at 2Hz, resulting in 65M video sequences for training.
For each sequence with 8 frames at 2Hz, we randomly take the leading $m\!\in\!\{1,2\}$ frames as conditional frames and the remaining $n\!\in\!\{7,6\}$ frames to be corrupted for video denoising, with probabilities $p\!\in\!\{0.1, 0.9\}$, respectively. 
We do not add noise on conditional frames since there is no need to generate \textit{past observations}.
The text condition is structured in the same way as the first stage, and we acquire the context from the middle frame of the sequence.
\modelname is trained on 64 GPUs for 112.5K iterations with a total batch size of 64. The learning rate is set as $1.25 \times 10^{-5}$ after $10^4$ warm-up steps.

In both stages, the input frames are resized to $256\times448$, and the text condition $\rvc$ is dropped
at a probability of $p = 0.1$ to enable classifier-free guidance~\cite{Ho2021classifierfree} in sampling. Both CLIP text encoders and the autoencoder are kept frozen throughout our experiments.

For extensions on action-conditioned prediction, we fine-tune the pre-trained \modelname as well as the linear projection layer for trajectory conditions on \nusdata. We conduct training on 16 GPUs for 100K steps with a total batch size of 16. Other training protocols such as the learning rate are the same with video prediction pre-training. For extensions on planning, we adapt a lightweight MLP to project the spatiotemporal features from frozen \modelname to future trajectory. We only optimize the MLP with 0.8M trainable parameters to adapt to planning.
The MLP is trained for 12 epochs with a batch size of 16 and a learning rate of $5\times10^{-4}$, taking only 10 minutes to converge on a single NVIDIA Tesla V100 device.

\subsection{Sampling Details}
\label{supp:sampling}
Given two types of conditions including the past two frames and text, \modelname simulates 6 future frames accordingly via iteratively denoising its input latent, which starts from random Gaussian noises.
The image resolution is $256\times448$ and the video sequence is at 2Hz.
The sampling process is performed by Denoising Diffusion Implicit Models (DDIM)~\cite{song2020DDIM}. We use 100 sampling steps and set the scale of classifier-free guidance to 7.5. The sampling speed is 539.41 ms/step.

\section{Experimental Setup}
\label{supp:exp_setup}

\subsection{Data Preparation}
We conduct extensive experiments on multiple datasets to evaluate the performance of our method. Specifically, the experiments of zero-shot transfer (\cref{supp:exp_unseen_gen}) are conducted on OpenDV-YouTube, Waymo~\cite{Sun2019WOD}, KITTI~\cite{Geiger2013KITTI} and Cityscapes~\cite{Cordts2016Cityscapes}. Experiments of action-condition prediction (\cref{supp:exp_world_model}) and motion planning~(Main Sec. 4.3) are established on \nusdata~\cite{Caesar2019nuScenes}. The results of text-to-image generation (\cref{supp:exp_img})
are shown in \youtubesplitename.
As for failure case studies (\cref{supp:failure_case}, \cref{fig:failure-case}), there are three cases in \youtubesplitename (a, b, d) and one case in Waymo (c).
All results are reported in the validation set, which is completely unseen in the training of \modelname. All images and video frames are resized to $256\times448$ before being fed into \modelname. For tasks based on video prediction, we construct 2 frames in 1s at 2Hz as conditional frames.
Each video sequence is paired with text conditions
composed of command and context.
For zero-shot datasets, the command and context are generated by the BLIP-2 model and video classifier respectively, following the preparation of training data.
For \nusdata, we generate the command from logged trajectory following~\cite{Hu2023UniAD, hu2022stp3} and map them to language using dictionary in~\cref{box:cmd}, and we take the scenario descriptions as the context, which are officially provided in the dataset.

\subsection{Metrics}
We use various metrics in multiple aspects for quantitative evaluation. These metrics include 
Fréchet Inception Distance (FID)~\cite{Heusel2017FID} , Fréchet Video Distance (FVD)~\cite{Unterthiner2018FVD}, 
CLIP-Similarity (CLIPSIM), Action Prediction Error, Average Displacement Error (ADE) and Final Displacement Error (FDE). For video prediction tasks, all predicted future frames are at 2Hz.
We refer readers for discussions on metrics in \cref{supp:discussion} (Q4).

\smallskip
\noindent\textbf{FID:} It evaluates the generation quality of images, which are video frames in our experiments, by measuring the distribution distance of features between the predictions and original frames in the dataset. The features are extracted by a pre-trained Inception model. For quantitative comparison on \nusdata, FID is evaluated on 6019 generated frames and ground-truth frames. For experiments on YouTube, FID is calculated on 18000 frames from both generation and the dataset.

\smallskip
\noindent\textbf{FVD:} It measures the semantic similarity between real and synthesized videos with a pre-trained I3D action classification model~\cite{Carreira2017I3D} as the feature extractor. We evaluate 4369 video clips for the \nusdata comparison experiment, and 3000 video clips for YouTube.

\smallskip
\noindent\textbf{CLIPSIM:} We use the CLIP  ViT-L/14~\cite{Radford2021CLIP} to 
evaluate the consistency and coherence of the predicted video by computing the average similarity score of CLIP features between 6 generated frames and the first conditional frame. We take 3000 video sequences for evaluation.

\smallskip
\noindent\textbf{Action Prediction Error:} 
For experiments of action-condition prediction on \nusdata, it measures the consistency between the input trajectory $\mathbf{w}$ and predicted future frames of \modelname. 
We transform the future frames into trajectory $\hat{\mathbf{w}}$ using an inverse dynamics model (IDM), which is trained on \nusdata to project a video sequence into a trajectory following the design in \cite{Lai2023xvo}.
This metric is then calculated as the mean L2 distance between all corresponding waypoints of $\mathbf{w}$ and $\hat{\mathbf{w}}$. 
Here both $\mathbf{w}$ and $\hat{\mathbf{w}}$ include 6 waypoints in 2 Hz, and $\mathbf{w}$ is generated from the logged trajectory in ego coordinate.

\smallskip
\noindent\textbf{ADE/FDE:} To evaluate the performance of planning on \nusdata, we calculate the ADE and FDE between the predicted trajectory and ground-truth trajectory in an open-loop setting.
Here, ADE is the mean L2 distance between all waypoints of these two trajectories, and FDE is the L2 distance between the final waypoints of them.

\section{More Visualizations}
\label{supp:exp_video_gen}

\subsection{Image Generation in Driving Domain}

\label{supp:exp_img}
\begin{figure*}[t!]
    \centering
    \includegraphics[width=\linewidth, height=0.9\textheight]{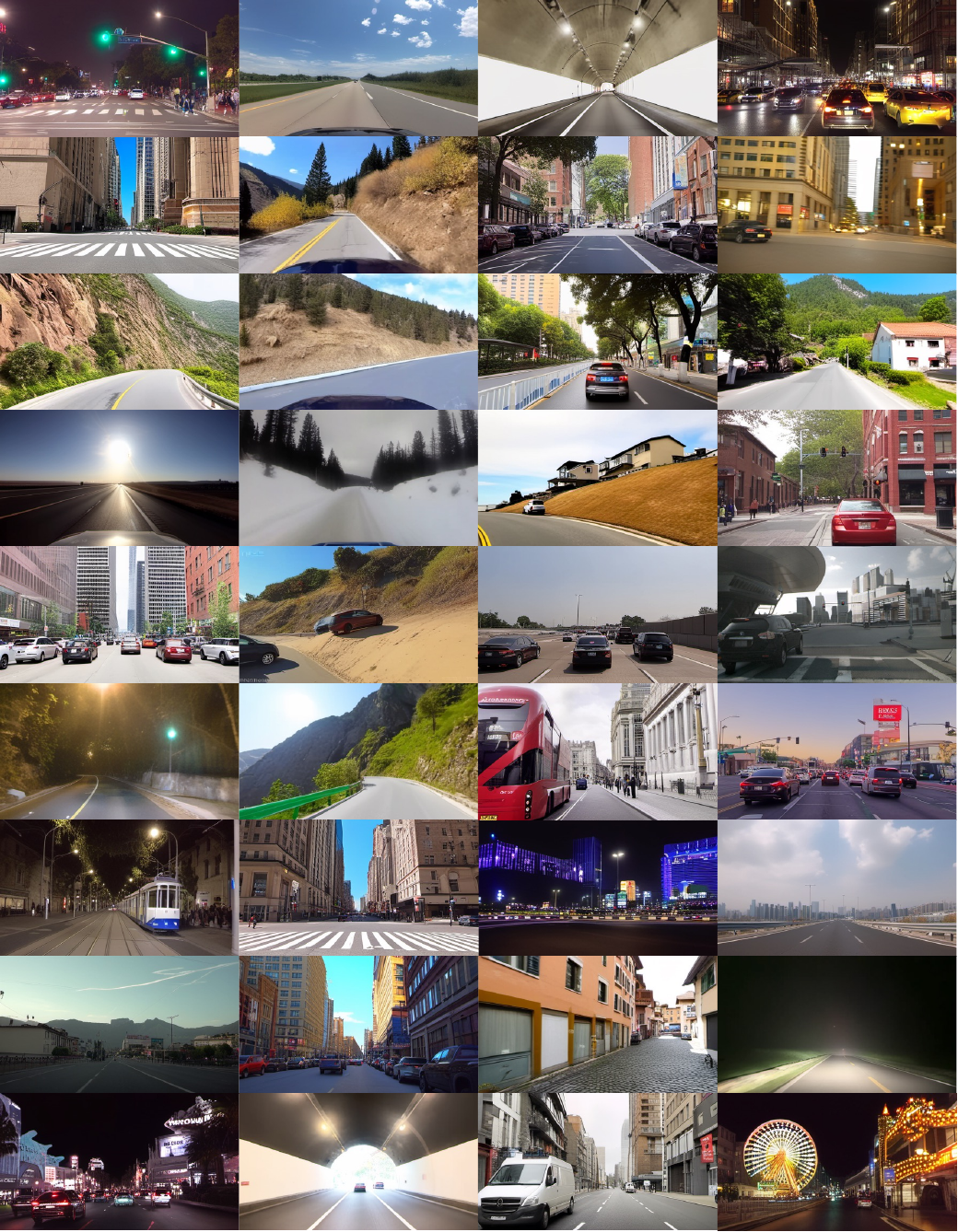}
    \vspace{5pt}
    \caption{\textbf{Generated images by the fine-tuned image model.} Corresponding text prompts are listed in \cref{box:image-gen-prompt}.
    }
    \label{fig:img}
\end{figure*}

After image domain transferring, the fine-tuned image model now focuses on synthesizing images in realistic driving views. Given text prompts in~\cref{box:image-gen-prompt}, the corresponding generated images are shown in~\cref{fig:img} where the generated samples greatly reflect the abundant visual details in complex and driving scenes. The ability of high-quality driving-view generation laid the foundation for simulating a realistic futuristic driving world, which is learned through video prediction pre-training.

\vspace{1pt}

{
\vspace{-1pt}
\captionsetup{type=table}
\begin{tcolorbox}[
                  colback=gray!10,%
                  colframe=black,%
                  width=\linewidth,
                  arc=1mm, auto outer arc,
                  boxrule=0.5pt,
                 ]
\scriptsize
\cmtt{1. Take a left turn. A city at night with a lot of lights. \\
2. Move steady. A car driving down a highway with a view of the sky. \\
3. Move steady. A car driving through a tunnel. \\
4. Drive steady. A city street at night with cars and taxis. \\
5. Keep the direction. A city street with a crosswalk and tall buildings. \\
6. Go straight. A car driving down a mountain road. \\
7. Maintain the direction. A city street with parked cars. \\
8. Turn to the left. A car driving down a city street. \\
9. Steer right. A car driving on a mountain road. \\
10. Make a right turn. A car driving down a mountain road. \\
11. Drive steady. A car driving down a city street. \\
12. Move steady. A car driving down a road in a small village. \\
13. Proceed. A car driving on a highway with a sun in the sky. \\
14. Drive steady. A car driving down a snowy road. \\
15. Take a left turn. A car driving down a hill with houses on the side. \\
16. Drive through the junction. A red car is driving down a street in Boston. \\
17. Brake. A city street with cars and tall buildings. \\
18. Proceed. A car is driving down a hill with parked cars on the side. \\
19. Decelerate. A car is driving on a highway with cars behind it. \\
20. Drive straight. \\
21. Move forward. \\
22. Move forward. A car driving on a mountain road. \\
23. Keep the direction. A red double decker bus driving down a city street. \\
24. Stop. A car driving on a busy street. \\
25. Drive straight. A tram on a street at night. \\
26. Drive forward. A city street with a crosswalk. \\
27. Proceed. A green light on a street with cars and pedestrians. \\
28. Move steady. A view of a highway with a city in the background. \\
29. Maintain the direction. A view of a city street with buildings and mountains in the background. \\
30. Drive straight. A city street with a lot of cars and buildings. \\
31. Steer right. A car driving down a cobblestone street in a city. \\
32. Drive forward. A car driving on a dark road at night. \\
33. Move forward. A car driving down a busy street at night. \\
34. Proceed. A car driving through a tunnel. \\
35. Drive straight. A white van driving down a city street. \\
36. Maintain the direction. A city street at night with a ferris wheel. \\
}
\end{tcolorbox}
\vspace{-10pt} 
\captionof{table}{\textbf{Prompts for image generation in \cref{fig:img}}, in the sequential order (from left to right and top to bottom).}
\label{box:image-gen-prompt}
}
\medskip

\subsection{Zero-shot Transfer}
\label{supp:exp_unseen_gen}
\begin{figure*}[t!]
    \centering
    \includegraphics[width=\linewidth]{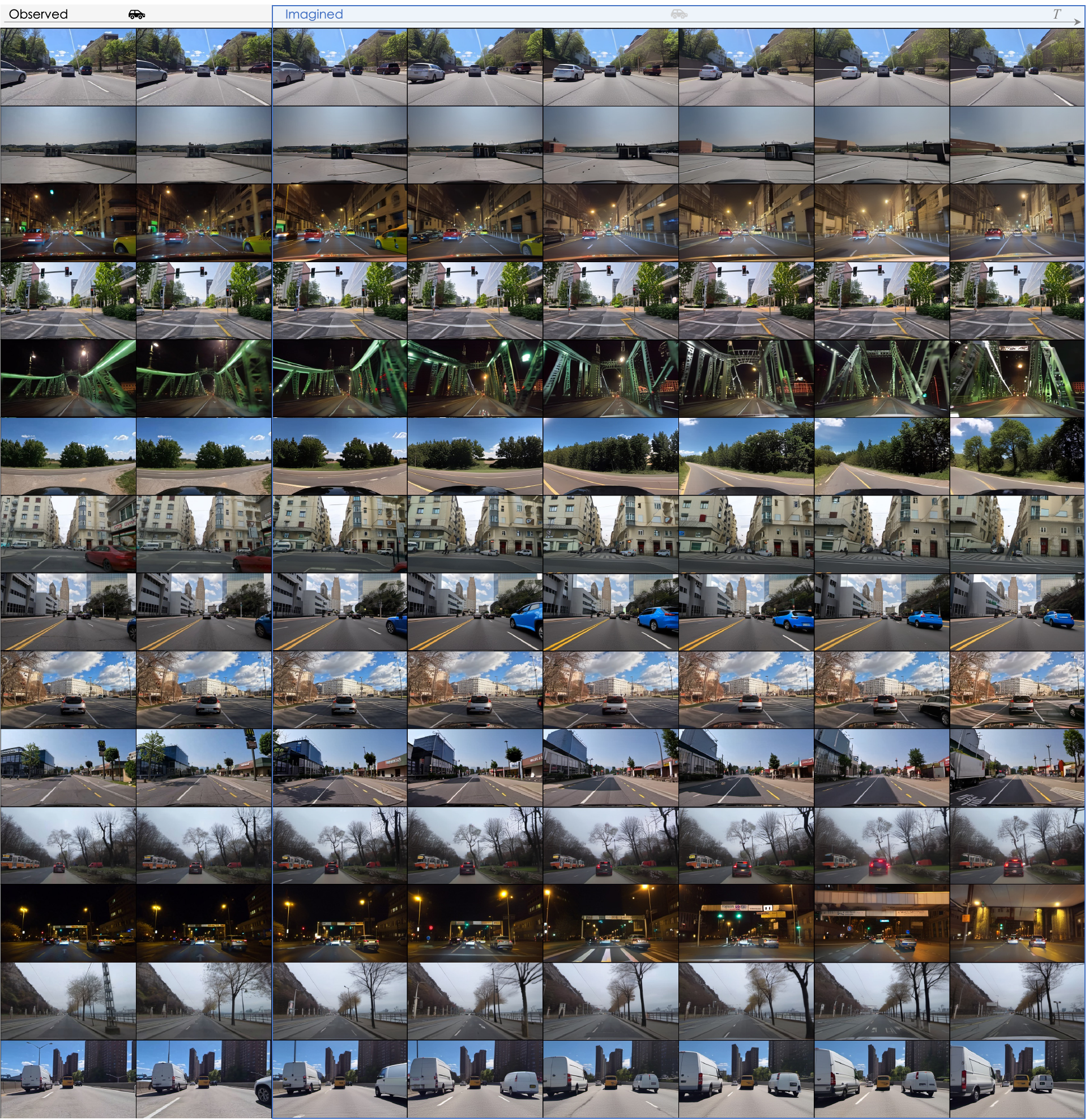}
    \caption{\textbf{Zero-shot video prediction on \youtubesplitename} (the YouTube-Val subset from different YouTubers with strict geofence). 
    The corresponding text conditions from top to bottom are as follows.
    1. ``Move steady. A car driving down a highway with cars behind it.",
    2. ``Turn to the left. A car is driving on a roof.",
    3. ``Maintain the direction. A taxi driving on a city street at night.",
    4. ``Drive forward. A car driving down a city street.",
    5. ``Proceed. A car driving on a bridge at night.",
    6. ``Steer left. A car driving on a road with trees and a blue sky.",
    7. ``Slow down. A street in a city with buildings and cars.",
    8. ``Go straight. A blue car driving down a city street.",
    9. ``Decelerate. A car driving on a city street.",
    10. ``Keep the direction. A view of a city street from the driver's seat.",
    11. ``Brake. A car driving down a street with trees and buses.",
    12. ``Proceed. A car driving on a city street at night.",
    13. ``Move forward. A car driving down a road near a river.",
    14. ``Drive straight. A van is driving down a highway with tall buildings in the background.".}
    \label{fig:zero-shot-youtube}
\end{figure*}

\begin{figure*}[t!]
    \centering
    \includegraphics[width=\linewidth]{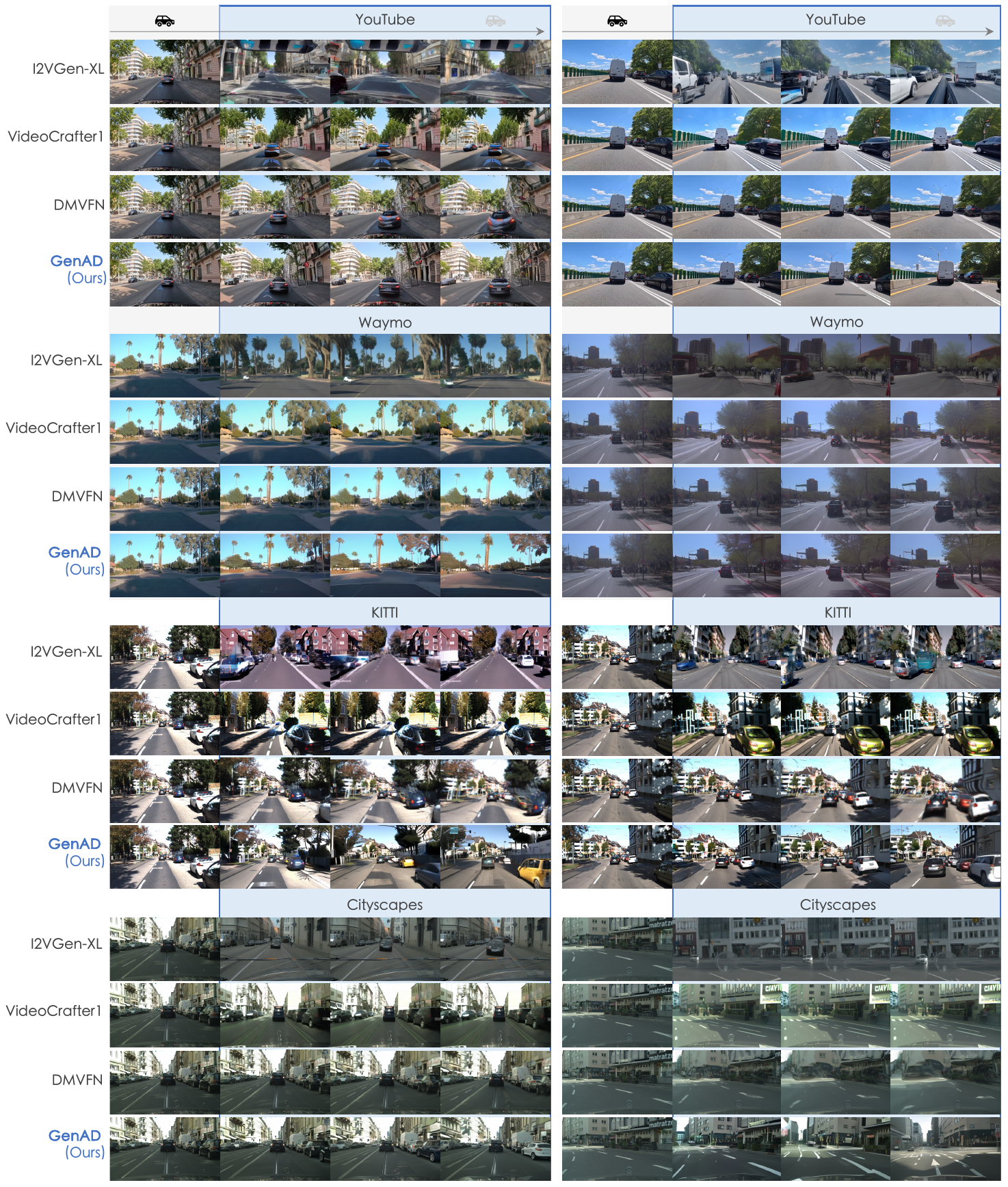}
    \caption{\textbf{Zero-shot video prediction on public datasets compared with state-of-the-art video generation/prediction models}. 
    Videos generated by I2VGen-XL are inconsistent with the condition frame. VideoCrafter1 appears to generate static scenarios. DMVFN suffers from huge image distortions. Meanwhile, all the other 3 models fail to generate videos when the ego vehicle should turn to the left and follow the lane (see the rightmost case in the last row). Our model manages to succeed in predictive video generation with great consistency with the conditional frames. We only show the first, third, and fifth frames from 6 predicted frames of our model due to space limits.   
    }
    \label{fig:zero-shot-public}
\end{figure*}

With a strong capability on video prediction, the pre-trained \modelname can generalize to multiple unseen datasets in a zero-shot manner. In \cref{fig:zero-shot-youtube}, we showcase multiple zero-shot video prediction results on \youtubesplitename.
In \cref{fig:zero-shot-public}, we illustrate the superiority of our method by comparing it to the previous state-of-the-arts on 4 datasets, including \youtubesplitename, Waymo~\cite{Sun2019WOD}, Cityscapes~\cite{Cordts2016Cityscapes} and KITTI~\cite{Geiger2013KITTI}. 

\begin{figure*}[t!]
    \centering
    \includegraphics[width=\linewidth]{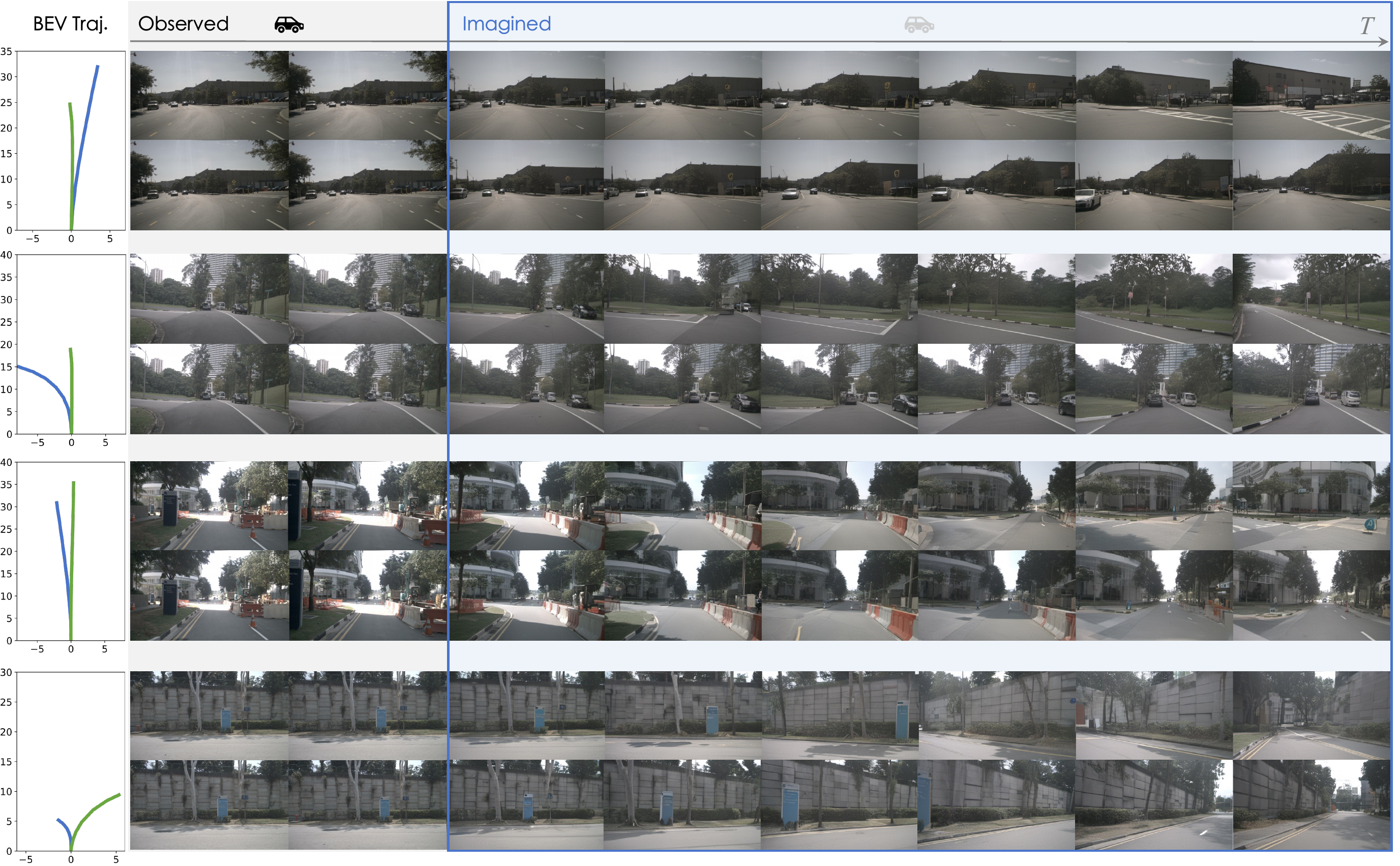}
    \vspace{-16pt}
    \caption{\textbf{Action-conditioned prediction on \nusdata}. We show four groups of video predictions for comparison, where each group is conditioned on the same two starting frames and different trajectories. In each group, the results in the first and second row are conditioned on the \textcolor{DarkBlue}{blue} and \textcolor{DarkGreen}{green} trajectories shown in the leftmost bird's-eye view, respectively.
    }
    \label{fig:action}
\end{figure*}

\subsection{Action-conditioned Prediction}
\label{supp:exp_world_model}

By introducing an additional trajectory condition, the fine-tuned \modelname-act can be controlled to simulate different futures according to the input trajectory. We show four groups of action-conditioned prediction in \cref{fig:action}. Both the input trajectory conditions (shown in the left bird’s-eye view map) and imagined future frames are in 3s at 2Hz.

\subsection{Failure Cases}
\label{supp:failure_case}

\begin{figure*}[b!]
    \centering
    \includegraphics[width=\linewidth]{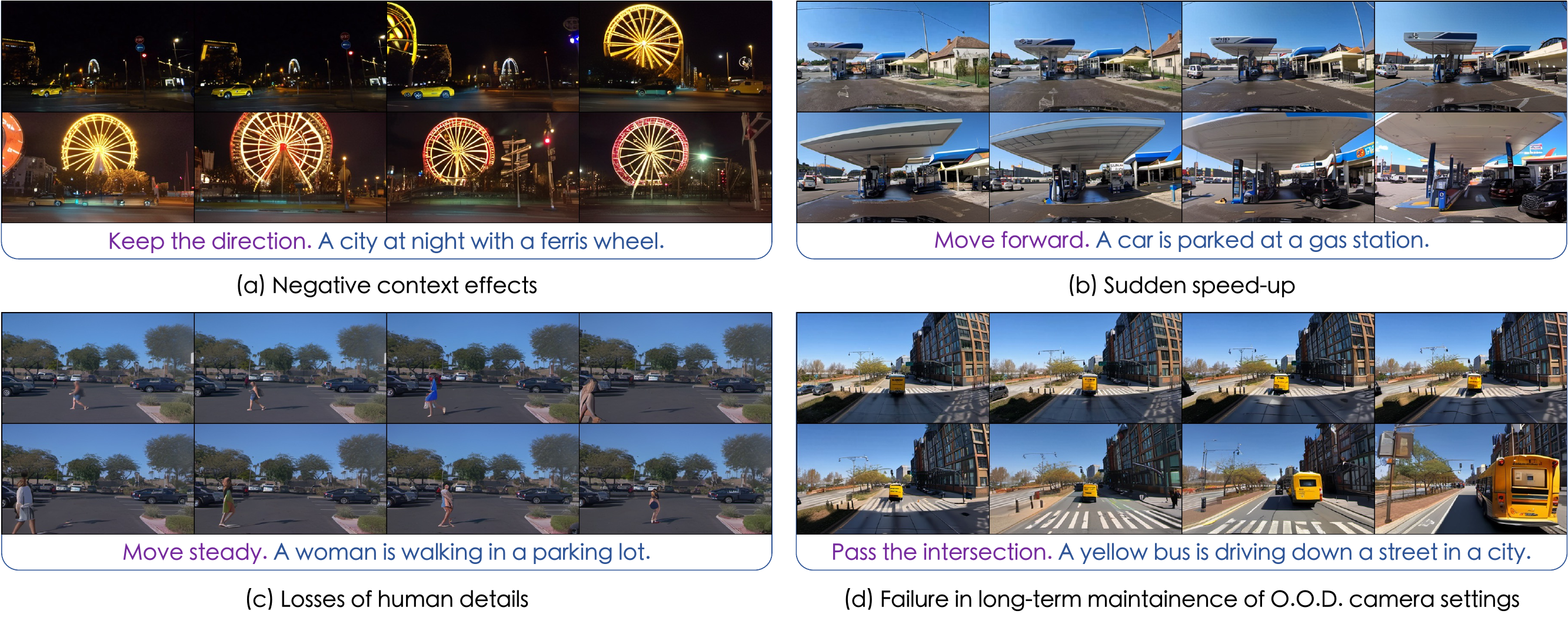}
    \vspace{-20pt}
    \caption{\textbf{Examples of failure cases}. Examples (a, b, d) are from \youtubesplitename, and example (c) is from Waymo. We notice that sometimes contexts exert negative impacts on generated videos since the model tends to sacrifice temporal consistency to explicitly generate the object in the context under some circumstances (see example (a)). In examples (b) and (c), the model faces challenges in generating smooth motion and human details, respectively. In example (d), the model succeeds in holding on to the out-of-distribution camera setting, \ie, on a double-deck bus, for the first 4 frames. But the camera height gradually falls down as normal in the last 2 frames.}
    \label{fig:failure-case}
\end{figure*}

We showcase four failure cases generated by our model in \cref{fig:failure-case}.
The model is sometimes disturbed by misleading contexts and is not strong enough to produce high-quality human details, as discussed in the \cref{supp:discussion} Q6. In some cases, the motion is not smooth enough. Meanwhile, the model fails to keep up with out-of-distribution camera height for 3s, even though succeeds in the first 2 seconds. 
These cases are worth future explorations.

\end{document}